\documentclass[11pt]{mystyle}

\usepackage[T1]{fontenc}
\usepackage{lmodern}
\usepackage{amsmath, amssymb, amsthm}
\usepackage{natbib}

\usepackage{fullpage}
\usepackage{wrapfig}
\usepackage{enumitem}
\setlist[itemize]{leftmargin=*,topsep=1pt,itemsep=1pt}
\usepackage{algorithm}
\usepackage{tikz}
\usetikzlibrary{positioning, fit, calc, shapes.geometric, arrows.meta, shadows, shadows.blur, backgrounds}

\usepackage{algpseudocode}

\usepackage{fancyhdr}
\newcommand{\projectpage}[1]{%
\begin{center}
\small
\vspace{-1.25ex}
\texttt{Project Page}: \url{#1}
\end{center}
}

\title{\textbf{Unlocking Feature Learning in Gated Delta Networks at Scale}}
\author{
    \textbf{Yifeng Liu}$^{1}$~~~~\textbf{Quanquan Gu}$^{2}$\\[1.5mm]
    $^1$University of California, Los Angeles\\[0.5mm]
    \texttt{\{liuyifeng,qgu\}@cs.ucla.edu}
}
\date{\today}

\begin{document}

\maketitle

\begin{abstract}
Training and scaling Large Language Models demand enormous computational resources, motivating both efficient sub-quadratic architectures and principled hyperparameter tuning methods. While the Maximal Update Parametrization ($\mu$P) has enabled zero-shot hyperparameter transfer for standard Transformers, its extension to linear models, particularly those with structured state transitions and complicated architectures, remains largely unexplored. By rigorously propagating coordinate-size estimates through the forward pass, gating mechanisms, and recurrent state dynamics, we derive the scaling rules for Gated Delta Network. Experiments on language-model pre-training confirm that our configurations enable stable learning-rate transfer across model widths under both AdamW and SGD, whereas standard parametrization fails to transfer, validating the correctness and practical utility of our analysis.
\end{abstract}

\section{Introduction}

The rapid development of Large Language Models (LLMs) has demonstrated remarkable capabilities across a wide range of downstream tasks~\citep{brown2020language,touvron2023llama, radford2019language,vaswani2017attention}. However, scaling these models to larger sizes introduces two challenges. First, empirical scaling laws show that optimal performance requires increased model size, while the computational budget required for training grows steeply with model scale~\citep{kaplan2020scaling,hoffmann2022training}. Second, the efficiency of standard Transformer architecture is limited by the quadratic complexity of softmax self-attention with respect to sequence length, making it increasingly costly for long-context inference and training~\citep{katharopoulos2020transformers}.

Linear models have been proposed to address these issues. The original linear attention~\citep{katharopoulos2020transformers} rewrites softmax attention as a linear kernel, enabling recurrent-form inference at constant per-step cost. Structured state space models (SSMs) such as S4~\citep{guefficiently}, Mamba~\citep{gu2024mamba} and Mamba-2~\citep{dao2024transformers} utilize recurrent state spaces to represent long-range dependencies within linear structures. A particularly promising family of linear recurrent models is based on the delta rule~\citep{widrow1960adaptive}, which updates a fast-weight matrix by subtracting the prediction error of the current key-value pair. Furthermore, the DeltaNet~\citep{yang2024deltanet} introduced a hardware-efficient parallel training algorithm for delta-rule Transformers, enabling scaling to large language models. Afterwards, Gated Delta Network~\citep{yang2025gated} augmented DeltaNet with the data-dependent gating mechanism of Mamba-2, which achieves strong language modeling performance while maintaining linear-time inference.
 
Simultaneously, training deep networks requires careful selection of hyperparameters such as learning rates, which are expensive to tune through grid search~\citep{snoek2012practical,snoek2015scalable}, and whose optimal values often change greatly with model scale. Meta-learning approaches have been explored to transfer hyperparameters across tasks and datasets~\citep{yogatama2014efficient,perrone2018scalable, horvath2021hyperparameter,akiba2019optuna}. A more principled solution is offered by the Maximal Update Parametrization ($\mu$P)~\citep{yang2020feature}, which identifies the valid parametrization of a neural network that supports feature learning in the infinite-width limit, as formalized through the Tensor Programs framework~\citep{yang2022tensor,yang2023tensor,TensorProgramVI}. $\mu$P theories demonstrated that hyperparameters tuned on small proxy models transfer zero-shot to large target models, with extensions to adaptive optimizers~\citep{yang2023tensor,ishikawaparameterization,everett2024scaling} and a spectral reformulation~\citep{yang2023spectral}. Subsequent work has successfully applied $\mu$P to other fields~\citep{blakeu2025,dey2024sparse,hajjar2024training}, and even industrial models~\citep{meta2024llama4,team2025longcat}.
 
Despite the development of efficient linear architectures, how to properly parametrize them for feature learning at scale has received very limited attention. The core challenge is that their recurrent state is updated through the sequence dimension, which does not fit the standard feedforward or attention-based derivations of $\mu$P. The only prior work on this is~\citet{vankadara2024feature}, which shows that vanilla $\mu$P and spectral scaling conditions both fail to support feature learning in diagonal SSMs like Mamba, and proposes a corrected scaling rule for them. However, the Gated Delta Network differs from diagonal SSMs fundamentally, since its recurrent state is a full matrix updated with additional data-dependent scalar gating through two separate weight matrices. These differences make the SSM-specific analysis of~\citet{vankadara2024feature} inapplicable, leaving the $\mu$P parametrization of Gated Delta Networks an open problem.
 
In this paper, we formally derive the complete $\mu$P formulation for Gated Delta Networks. Our main contributions are:
\begin{itemize}
  \item  We theoretically derive coordinate-size estimates through the full forward pass. We also derive principled initialization variances, forward multipliers, and learning-rate scalings for all weight classes. We discover that the gating weight matrices require a non-standard $\Theta(1/\sqrt{d})$ learning-rate scaling, and the scalar gating parameters require a $\Theta(\sqrt{d})$ scaling, both of which deviate from standard $\mu$P setting.
 
  \item We pretrain Gated Delta Network language models across multiple widths and show that our $\mu$P formulation enables zero-shot learning-rate transfer under both AdamW and SGD optimizers, while standard parametrization fails to transfer, confirming both the theoretical derivation and its practical efficiency.
\end{itemize}

\section{Related Work}
 
\textbf{Efficient Sequence Models} The standard Transformer~\citep{vaswani2017attention} and its variants~\citep{radford2019language, brown2020language,touvron2023llama} have become the dominant architecture for large-scale language modeling, but their quadratic attention complexity limits its application in much larger language models. Linear attention~\citep{katharopoulos2020transformers} replaces the softmax with a kernel function that permits rewriting attention as a linear RNN, enabling $O(1)$ per-step inference. Structured state space models (SSMs) improve by integrating recurrent state spaces. S4~\citep{guefficiently} introduces HiPPO-based~\citep{gu2020hippo} structured matrices for long-range sequence modeling, Mamba~\citep{gu2024mamba} adds input-selective state transitions for improved performance, and Mamba-2~\citep{dao2024transformers} unifies SSMs with structured matrix attention via state-space duality. Other notable architectures include RetNet~\citep{sun2023retentive},  RWKV~\citep{peng2023rwkv}, Gated Linear Attention~\citep{yang2024gla},  HGRN~\citep{qin2023hgrn} and its expansion HGRN2~\citep{qin2024hgrn2}. Recently, numerous models consider applying delta rule~\citep{widrow1960adaptive}, which subtracts the prediction error rather than accumulating outer products. This was formalized in the fast-weight programmer framework~\citep{schlag2021linear}, with recurrent extensions in~\citet{irie2021going}. DeltaNet~\citep{yang2024deltanet} further introduced a hardware-efficient parallel training algorithm, and Gated Delta Networks~\citep{yang2025gated} combined the delta-rule state update with the data-dependent gating of Mamba-2 with improved performances. Our work focuses precisely on this architecture and derives its $\mu$P formulation.
 
\textbf{Hyperparameter Transfer} A lot of researchers have explored approaches to accelerate the search progress of hyperparameters of training deep learning models~\citep{snoek2012practical,snoek2015scalable,jamieson2016non,akiba2019optuna}. Some have also explored the methods to transfer learning between different tasks or datasets~\citep{horvath2021hyperparameter,perrone2018scalable,yogatama2014efficient}. Moreover, based on standard parametrization (SP, such as Xavier initialization~\citep{glorot2010understanding} and Kaiming initialization~\citep{he2015delving}), \citet{yang2020feature} proposed Maximal Update Parametrization ($\mu$P) based on abc-parameterization framework, which unifies previous parametrization methods such as SP, Neural Tangent Kernel (NTK)~\citep{jacot2018neural} and Mean Field parametrization~\citep{chizat2018global,mei2018mean,sirignano2020mean,rotskoff2022trainability}. It enables feature learning that can be generalized to infinite-width conditions. And based on $\mu$P,~\citet{yang2022tensor} proposed $\mu$\textit{Transfer} that can implement zero-shot transfer of hyperparameters to large models from a much smaller proxy model, and generalized it to different architectures and optimizers~\citep{yang2023tensor}, such as SGD, Adagrad~\citep{duchi2011adaptive} and Adam~\citep{adam}. Following that, they also reformulated $\mu$P from the perspective of spectral norm~\citep{yang2023spectral}. Recently, many researchers tried to further generalize $\mu$P in other fields or successfully scale LLMs with $\mu$P ~\citep{blakeu2025,meta2024llama4,haas2024effective,dey2024sparse,hajjar2024training,team2025longcat}.

However, the question of how to properly parametrize these models for feature learning at scale has received very little attention. The core challenge is that original $\mu$P designs cannot be directly applied to these architectures with recurrent state transitions and structured matrix operations. To our knowledge, the only work that formally addresses $\mu$P-style parametrization for this class of models is~\citet{vankadara2024feature}, which studies the scaling behavior of structured SSMs like Mamba. Their analysis reveals that both vanilla $\mu$P and spectral scaling conditions fail to support feature learning in SSMs, and they derive the scaling rule for SSMs that recovers feature learning. The parametrization of Gated Delta Net, however, differs from the diagonal SSMs greatly, since its state is updated via outer-product delta rules rather than scalar recurrences with matrix-valued hidden states. This work is, to our knowledge, the first to derive and validate a $\mu$P-consistent parametrization for Gated Delta Network.

\section{Preliminaries}
\subsection{Gated Delta Net}
Proposed by~\citet{yang2025gated}, Gated Delta Net is a variant of linear transformer~\citep{katharopoulos2020transformers}, based on the Mamba 2 architecture~\citep{dao2024transformers}. For the query, key and value vectors $\qb_t,\kb_t$ and $\vb_t$ similar to the original Transformer, the update rule of the latent state is shown as:
\begin{align}
    \Sbb_t=\Sbb_{t-1}(\alpha_t(\Ib-\beta_t\kb_t\kb_t^\top))+\beta_t\vb_t\kb_t^\top,\label{eq:update_rule}
\end{align}
where $\alpha_t\in(0,1)$ is the data-dependent gating scale and $\beta_t\in(0,1)$ is the ``writing strength'' of the current input at time $t$, as proposed in~\citet{widrow1960adaptive,schlag2021linear}. And the output is just direct readout of the latent state on the query:
\begin{align}
    \ob_t=\Sbb_t\qb_t.~\label{eq:readout}
\end{align}
Different from the Transformer, Gated Delta Net added a short convolution after query, key and value projections, followed by a SiLU activation layer. There are also L2 Normalization layer for queries and keys. And there is also an RMSNorm layer before the output projection to stabilize the training. As discussed in the original paper, these norm are crucial to the performance of Gated Delta Net.
\subsection{$\mu$P theory}
In deep learning, models are frequently scaled by increasing their hidden dimension or width $d$. Under the Standard Parameterization (SP), including He~\citep{he2015delving} or Xavier~\citep{glorot2010understanding} initialization, hidden weights are typically initialized with entries drawn from $\mathcal{N}(0, \sigma^2/d)$ and optimized using a uniform learning rate $\eta$ across all layers. However, as $d$ goes to infinity, SP encounters fundamental limitations. If the learning rate remains constant, the network's activations and gradients diverge. To prevent this instability, $\eta$ must be scaled down by $\mathcal{O}(1/d)$, which forces the network into the Neural Tangent Kernel (NTK) or ``lazy training'' regime \cite{yang2020feature}, where the intermediate representations (features) seldom evolve from their initialized state, meaning the network fails to perform real feature learning.

To resolve the trade-off between stability and feature learning in the infinite-width limit, \citet{yang2020feature} proposed the \textit{Maximal Update Parameterization} ($\mu$P) using the Tensor Programs framework. $\mu$P provides rigorous configurations for scaling weight initializations and learning rates as a function of the width $d$ (sometimes a width-dependent multiplier on the weight is required; refer to Tables~\ref{tab:gdn_formulation} and~\ref{tab:gdn_sgd_formulation} for AdamW and SGD configurations) to ensure feature learning. In this setting, feature updates at every layer remain bounded and non-vanishing (i.e., $\Delta h = \Theta(1)$) as the model expands to infinity width. To further illustrate this, the definition of coordinate size should be first introduced:
\begin{definition}
A vector $v\in \RR^d$ has $\Theta(d^a)$-sized coordinates if $\|v\|^2/d=\Theta(d^{2a})$, i.e., each entry of $v$ has variance $\Theta(d^{2a})$ as $d\to\infty$. When $d$ is large, the coordinates of the vectors being studied are regarded as roughly i.i.d. Gaussian.
\end{definition}

Based on the definition above, $\mu$P theory proposes three desiderata. Firstly, every (pre)activation vector should have $\Theta(1)$-sized coordinates; and the output of a network should be $O(1)$; moreover, all parameters should be updated as much as possible without leading to divergence. And based on these desiderata and the assumption of feature learning, there are some derivatives. For example, the gradient to a hidden state is with $\Theta(1/d)$ coordinate size when optimized with SGD optimizer.

\section{The $\mu$P Forward Analysis of Gated Delta Net}
In this section, we will review the architecture of Gated Delta Net, and then derive the scaling law of this architecture.
\label{sec:mup_analysis}

We derive the Maximal Update Parametrization ($\mu$P) conditions for Gated Delta Net by propagating coordinate-size estimates through the forward pass and the gating mechanisms, then conclude with the implications for the AdamW optimizer.

\paragraph{Notation and standing assumptions.}
Following~\citet{yang2022tensor}, we say a vector $\zb \in \RR^d$ has $\Theta(1)$ coordinate size if $\|\zb\|_2 = \Theta(\sqrt{d})$, i.e.\ each coordinate is of order $\Theta(1)$ in magnitude. Equivalently, the per-coordinate variance of $\zb$ is $\Theta(1)$. For a matrix $\Ab \in \RR^{d \times d}$, we say it has $\Theta(c)$ coordinate size if each entry is of order $\Theta(c)$ in magnitude.

We assume throughout that the hidden state $\xb_t \in \RR^d$ satisfies the $\mu$P feature-learning condition, namely
\begin{align*}
    \|\xb_t\|_2 = \Theta(\sqrt{d}), \qquad
    \|\Delta \xb_t\|_2 = \Theta(\sqrt{d}),
\end{align*}
so that $\xb_t$ has $\Theta(1)$ coordinate size and its update is of the same order. To isolate the effect of the parametrization we temporarily ignore the SiLU activations (see Remark~\ref{rem:silu} below).

\subsection{Coordinate sizes of the projected features}
Let $\tilde{\qb}_t = \text{ShortConv}(\Wb_q\,\xb_t)$, $\tilde{\kb}_t = \text{ShortConv}(\Wb_k\,\xb_t)$, and $\vb_t = \text{ShortConv}(\Wb_v\,\xb_t)$, where $\Wb_q, \Wb_k, \Wb_v \in \RR^{d \times d}$ are the query, key, and value projection matrices, respectively. Under the $\mu$P initialization of hidden weights~\citep{yang2022tensor}, the products $\Wb_q\,\xb_t$, $\Wb_k\,\xb_t$, $\Wb_v\,\xb_t$ each have $\Theta(1)$ coordinate size; the short convolution preserves this order, so $\tilde{\qb}_t, \tilde{\kb}_t$ and $\vb_t$ each have $\Theta(1)$ coordinate size. The L2-normalized query and key are
\[
    \qb_t = \frac{\tilde{\qb}_t}{\|\tilde{\qb}_t\|_2}, ~
    \kb_t = \frac{\tilde{\kb}_t}{\|\tilde{\kb}_t\|_2}.
\]
Since $\|\tilde{\qb}_t\|_2 = \Theta(\sqrt{d})$, each coordinate of $\qb_t$ and $\kb_t$ is of order $\Theta(1)/\Theta(\sqrt{d})=\Theta(1/\sqrt{d})$, i.e., $\qb_t$ and $\kb_t$ both have $\Theta(1/\sqrt{d})$ coordinate size. 

\subsection{Coordinate size of the latent state}~\label{coordinate_size_of_latent_state}
The rank-one write update in~\eqref{eq:update_rule} is $\Ub_t = \beta_t\,\vb_t\kb_t^{\top}$. Since $\beta_t \in (0,1)$ as a bounded scalar and combining the $\Theta(1)$ coordinate size of $\vb_t$ with the $\Theta(1/\sqrt{d})$ coordinate size of $\kb_t$, each entry of $\Ub_t$ satisfies
\[
    (\Ub_t)_{ij}=\beta_t\,(\vb_t)_i\,(\kb_t)_j=\Theta(1) \cdot \Theta\Big(\frac{1}{\sqrt{d}}\Big)=\Theta\Big(\frac{1}{\sqrt{d}}\Big).
\]
For the cumulative latent state $\Sbb_t$, we apply the argument of~\citet{vankadara2024feature}\footnote{The detailed derivations can be found in Appendix~\ref{sec:cumulative_latent_space}.}: unless the write update $\Ub_t$ perfectly cancels the residual term in~\eqref{eq:update_rule} at every step $t$, the steady-state variance of $\Sbb_t$ matches that of $\Ub_t$. More precisely, the spectral contraction factor of the map $\Sbb \mapsto \Sbb\,\alpha_t(\Ib - \beta_t\kb_t\kb_t^{\top})$ is at most $\alpha_t(1 - \beta_t\|\kb_t\|^2_2) \leq \alpha_t$, which is strictly less than $1$ when $\alpha_t,\beta_t\in(0,1)$ are both bounded away from $0$ and $1$. We assume this condition holds throughout the analysis. Under this assumption the geometric sum of write updates converges and $\Sbb_t$ has $\Theta(1/\sqrt{d})$ coordinate size.

\subsection{Coordinate size of the readout}
The output is $\ob_t = \Sbb_t\,\qb_t$, so the $j$-th coordinate is \[(\ob_t)_j = \sum_{i=1}^{d} (\Sbb_t)_{ji}\,(\qb_t)_i.\] Treating the entries of $\Sbb_t$ and $\qb_t$ as approximately independent and zero-mean, each with variance $\Theta(1/d)$, the variance of the sum is
\[
    \text{Var}\bigl[(\ob_t)_j\bigr]
    = \sum_{i=1}^{d}
      \text{Var}\bigl[(\Sbb_t)_{ji}\bigr]\cdot
      \text{Var}\bigl[(\qb_t)_i\bigr]
    = d \cdot \Theta\Big(\frac{1}{d}\Big)\cdot\Theta\Big(\frac{1}{d}\Big)= \Theta\Big(\frac{1}{d}\Big),
\]
so $\ob_t$ has $\Theta(1/\sqrt{d})$ coordinate size. Although the subsequent RMSNorm forces its output to $\Theta(1)$ coordinate size, this implicit rescaling would disrupt the gradient scaling required by $\mu$P. We therefore recommend inserting a $\sqrt{d}$-multiplier before RMSNorm so that the input to RMSNorm is already $\Theta(1)$:
\[
    \ob_t \mapsto \sqrt{d}\ob_t,
    \qquad \|\sqrt{d}\ob_t\|_2 = \Theta(\sqrt{d}).
\]
An equivalent alternative is to replace the L2-Normalization on $\qb_t$ with an RMSNorm layer, which absorbs the same $\sqrt{d}$-factor. With either modification, the standard $\mu$P formulation applies to all projection weights other than those governing $\alpha_t$ and $\beta_t$.

\subsection{First-order analysis of the gating scalars}

The gating parameters are defined as
\[\beta_t = \sigma(\Wb_\beta\,\xb_t) \in (0,1),\qquad \Wb_\beta \in \RR^{1 \times d},\]
and
\[\alpha_t = e^{g_t} \in (0,1), \qquad g_t = -e^{a_{\log}}\ln(1 + e^{\,\Wb_\alpha\xb_t + b}),
\]
with $\Wb_\alpha \in \RR^{1 \times d}$ a trainable weight row and $a_{\log}, b \in \RR$ scalar parameters shared within each head. Because both $\alpha_t$ and $\beta_t$ are nonlinear transformations of a Gaussian-distributed pre-activation, they are not themselves Gaussian, and traditional $\mu$P theory does not directly apply. 

For $\alpha_t$, since it is bounded by $(0,1)$, it is naturally $\Theta(1)$. Define $z_{\alpha,t}=\Wb_\alpha \xb_t+b$, when $z_{\alpha,t} + a_{\log}\ll 0$, $|\partial\alpha_t/\partial z_{\alpha,t}| = \alpha_t e^{z_{\alpha,t}+a_{\log}}/(1+e^{z_{\alpha,t}})$ is also $\Theta(1)$. Under the original initialization of~\citet{yang2025gated}, namely $a_{\log} \sim \mathrm{Uniform}(0,16)$ and $b = b_0 + \ln(1 - e^{-b_0})$ with $b_0 = 10^{2\epsilon_b - 3}$, $\epsilon_b \sim \mathrm{Uniform}(0,1)$, the gradient $|\partial\alpha_t/\partial z_{\alpha,t}| = \alpha_t\cdot e^{z_{\alpha,t}+a_{\log}}/(1+e^{z_{\alpha,t}})=e^{z_{\alpha,t}+a_{\log}}/(1+e^{z_{\alpha,t}})^{e^{a_{\log}}+1}$ is bounded by $1$. 
\begin{proof}
Denote $k=e^{a_{\log}}>0,p=e^{z_{\alpha,t}}>0$, and $f(k,p)=\log(\frac{kp}{{(1+p)}^{k+1}})=\log(kp)-(k+1)\log(1+p)$. Then we have $\frac{\partial f}{\partial {p}}=\frac{1}{p}-\frac{1+k}{1+p}=\frac{1-pk}{p(1+p)}$. Therefore, $\forall k>0$, $f(k,p)\le f(k,1/k)=\log(\frac{1}{(1+1/k)^{k+1}})<0$, and $|\partial\alpha_t/\partial z_{\alpha,t}|=e^{f(k,p)}<1$. 
\end{proof}

For $\beta_t$, since $z_{\beta,t} := \Wb_\beta\xb_t$ is $\Theta(1)$, $\beta_t\in (0,1)$ does not saturate and $\beta_t=\sigma(z_{\beta,t})$, therefore, $\frac{\partial\beta_t}{\partial z_{\beta,t}}=\beta_t(1-\beta_t)=\Theta(1)$.

Combining the two analyses, $\Theta(1)$ coordinate-size behavior of both $\alpha_t$ and $\beta_t$ is maintained under $\mu$P initialization. Consequently, $\Wb_\alpha$ and $\Wb_\beta$ may be treated as hidden weights with initial variance $1/d$, while the scalar parameters $a_{\log}$ and $b$ are assigned constant (i.e., width-independent) initial variance.

\begin{remark}[Effect of SiLU activations]
\label{rem:silu}
The analysis above assumes that the SiLU activations $\text{SiLU}(x) = x/(1+e^{-x})$ following the short convolutions are suppressed. In practice, these activations introduce non-Gaussian statistics in $\tilde{\qb}_t$, $\tilde{\kb}_t$, and $\vb_t$. Therefore, the derivations above are only an approximation to the ideal $\mu$P conditions. The approximation quality degrades if the pre-activations are far from zero, but remains adequate for the initialization in $\mu$P.
\end{remark}

\section{The $\mu$P Analysis of Gated Delta Net under SGD}
\label{sec:mup_sgd}

In this section, we will derive the backward process of the $\mu$P configuration and the learning rates for each module accordingly. In this section, we focus on the scenario of Gated Delta Net under SGD and postpone the analysis for AdamW to Appendix~\ref{sec:compat_adamw}, where AdamW optimizer benefits from a key simplification: it normalizes the gradient by its coordinate-wise second moment, so the effective update magnitude is $\Theta(\eta)$ per coordinate regardless of the raw gradient scale.  Consequently, all weight classes share the same $\Theta(1)$ update magnitude, and the learning-rate schedule needs only to account for the number of terms in the activation sum (i.e.\ the fan-in $n_{\ell-1}$).

Under plain SGD this normalization is absent. The update rule is
\begin{align*}
    W^\ell \leftarrow W^\ell - \eta^\ell \nabla_{W^\ell}\mathcal{L},
\end{align*}
so the magnitude of the change $\Delta z^\ell = (\Delta W^\ell) h^{\ell-1}$ is proportional to both the learning rate and the raw gradient magnitude. Different weights in Gated Delta Net receive different gradient magnitudes, so they require different learning-rate scalings to satisfy the $\mu$P feature-learning condition $\|\Delta h^\ell\|_2 = \Theta(\sqrt{d})$.

\subsection{Notation and assumptions} 

We retain all conventions and assumptions from Section~\ref{sec:mup_analysis}: the hidden state $\xb_t$ satisfies $\|\xb_t\|_2 = \Theta(\sqrt{d})$, all forward-pass estimates carry over unchanged, and SiLU activations are suppressed (see Remark~\ref{rem:silu}). And we write $\eta^\ell$ for the per-layer SGD learning rate.

\begin{assumption}[Short effective memory and domination of direct gradient]
\label{ass:short_memory}
The loss gradient with respect to the latent state satisfies
\begin{align*}
    \frac{\partial\mathcal{L}}{\partial\Sbb_t}
    = \gb_t\qb_t^{\top}
    + \underbrace{\sum_{\tau=t+1}^{T}
      \Bigl(\prod_{s=t+1}^{\tau}\alpha_s\Bigr)
      \gb_\tau\qb_\tau^{\top}
      \Bigl(\prod_{s=t+1}^{\tau}\bigl(\Ib-\beta_s\kb_s\kb_s^{\top}\bigr)\Bigr)}_{\text{BPTT tail}},
\end{align*}
where the first term is the direct contribution from the readout at step $t$ and the second term accumulates gradients from all future readouts through the recurrent state chain. Each BPTT (Backpropagation-through-time) term is individually of the same $d$-order as the direct term: its $(a,b)$-entry has magnitude $\Theta(1/d)$, identical to $(\gb_t)_a(\qb_t)_b$. However, the BPTT sum contains up to $T-t$ terms.

We assume throughout this section that the effective memory length $L_{\mathrm{eff}} := \sum_{\tau=t}^{T}\prod_{s=t+1}^{\tau}\alpha_s = O(1)$, i.e.\ the gating values $\alpha_s$ decay the past state sufficiently fast so that $\bar\alpha := \mathbb{E}[\alpha_t]$ satisfies $(1-\bar\alpha)^{-1}=O(1)$. Under this assumption the BPTT tail is $O(1)$ times the direct term in the $\Theta(\cdot)$ sense, and all gradient estimates below use only the direct term $\partial\mathcal{L}/\partial\Sbb_t = \gb_t\qb_t^{\top}$ without loss of $d$-scaling accuracy.

When the model is used in a long-context scenario with slow forgetting ($\alpha_t\to 1$), $L_{\mathrm{eff}}$ can grow as $O(T)$. In this case, the $\mu$P learning-rate prescriptions should be scaled down by $1/L_{\mathrm{eff}}$ accordingly. And we postpone the detailed derivations of the BPTT term in Appendix~\ref{sec:additional_backward_derivation}.
\end{assumption}

\subsection{Backward error at the readout} 
The readout is $\ob_t = \Sbb_t \qb_t$, followed by a $\sqrt{d}$-multiplier and RMSNorm. The loss-gradient at $\ob_t$ satisfies
\[\gb_t:= \frac{\partial\mathcal{L}}{\partial\ob_t}= \sqrt{d}\cdot\frac{\partial\mathcal{L}}{\partial(\sqrt{d}\,\ob_t)}.
\]
If we assume that all the architectures outside Gated Delta Net follow $\mu$P rules, then it is assumed that $\frac{\partial\mathcal{L}}{\partial(\sqrt{d}\,\ob_t)}$ has $\Theta(1/d)$ coordinate size and $\gb_t$ is with $\Theta(1/\sqrt{d})$ coordinate size. 

\subsection{Gradient of query, key and value projections}
\paragraph{Gradient of the value projection $\Wb_v$.}
Recall from~\eqref{eq:update_rule} and~\eqref{eq:readout} that $\Sbb_t$ depends on $\vb_t$ through the rank-one write update $\beta_t\vb_t\kb_t^{\top}$. Under Assumption~\ref{ass:short_memory}, the gradient of $\mathcal{L}$ with respect to $(\vb_t)_i$ receives a contribution from the direct readout at time $t$:
\begin{align}
    \frac{\partial\mathcal{L}}{\partial(\vb_t)_i}
    &= \beta_t\sum_j
       \frac{\partial\mathcal{L}}{\partial(\Sbb_t)_{ij}}
       \cdot(\kb_t)_j = \beta_t\sum_j (\gb_t)_i (\qb_t)_j \cdot (\kb_t)_j
    = \beta_t\,(\gb_t)_i\,\langle\qb_t,\kb_t\rangle,
    \label{eq:grad_v}
\end{align}
where we used $\partial\mathcal{L}/\partial(\Sbb_t)_{ij} = (\gb_t)_i(\qb_t)_j$ from the readout~\eqref{eq:readout} under the short-memory assumption (Assumption~\ref{ass:short_memory}). Note also that $\vb_t$ does not propagate gradient back to earlier states $\Sbb_{t'}$ with $t'<t$; those states depend on $\vb_{t'}$, not $\vb_t$. However, $\Sbb_t$ itself contributes to future states $\Sbb_{t''}$ for $t''>t$, which is precisely the BPTT tail handled by Assumption~\ref{ass:short_memory}.

Since $\qb_t$ and $\kb_t$ are L2-normalized to unit vectors in $\RR^d$, with independent entries each of order $\Theta(1/\sqrt{d})$ at initialization, the inner product satisfies $\bigl|\langle\qb_t,\kb_t\rangle\bigr|= \Theta\big(\frac{1}{\sqrt{d}}\big)$.
In~\eqref{eq:grad_v} the three factors are: $\beta_t = \Theta(1)$, $(\gb_t)_i = \Theta(1/\sqrt{d})$ per coordinate, and $\langle\qb_t,\kb_t\rangle = \Theta(1/\sqrt{d})$. Hence:
\begin{align*}
    \frac{\partial\mathcal{L}}{\partial(\vb_t)_i}
    = \Theta(1)\cdot\Theta\Big(\frac{1}{\sqrt{d}}\Big)\cdot\Theta\Big(\frac{1}{\sqrt{d}}\Big)
    = \Theta\Big(\frac{1}{d}\Big).
\end{align*}
By treating ShortConv as the identity for scale estimates, the gradient of the value projection $\Wb_v \in \RR^{d\times d}$ is
\[
    \frac{\partial\mathcal{L}}{\partial(\Wb_v)_{ij}} = \frac{\partial\mathcal{L}}{\partial(\vb_t)_i}\cdot(\xb_t)_j = \Theta\Big(\frac{1}{d}\Big)\cdot\Theta(1) = \Theta\Big(\frac{1}{d}\Big).
\]
The SGD update is $\Delta(\Wb_v)_{ij} = -\eta_v\Theta(1/d)$.
The resulting change in $\vb_t$ is
\[
    (\Delta\vb_t)_i = \sum_{j=1}^d \Delta(\Wb_v)_{ij}(\xb_t)_j = d\cdot\Theta\Big(\frac{\eta_v}{d}\Big) = \Theta(\eta_v).
\]
The feature-learning condition requires $\Delta\vb_t$ with $\Theta(1)$ coordinate size, therefore, 
\begin{align*}
    \boxed{\eta_v = \Theta(1).}
\end{align*}

\paragraph{Gradient of the key projection $\Wb_k$.}
The key $\kb_t = \tilde{\kb}_t/\|\tilde{\kb}_t\|_2$ enters the latent-state update both through the write term ($\beta_t\vb_t\kb_t^{\top}$) and the erase term ($\alpha_t\Sbb_{t-1}(\Ib - \beta_t\kb_t\kb_t^{\top})$). The gradient of $\mathcal{L}$ with respect to the normalized $\kb_t$ from the write term alone is
\begin{align}
    \Big(\frac{\partial\mathcal{L}}{\partial\kb_t}\Big)_l\Bigg|_{\text{write}}
    = \beta_t\sum_i\frac{\partial\mathcal{L}}{\partial(\Sbb_t)_{il}}\cdot(\vb_t)_i= \beta_t\,(\qb_t)_l\sum_i(\gb_t)_i(\vb_t)_i= \beta_t\,(\qb_t)_l\,\langle\gb_t,\vb_t\rangle.
    \label{eq:grad_k_write}
\end{align}
Here $\vb_t$ has $\Theta(1)$ coordinate size while $\gb_t$ has $\Theta(1/\sqrt{d})$ coordinate size, so $|\langle\gb_t,\vb_t\rangle|= \Theta(1)$. Since $\qb_t$ and $\kb_t$ have $\Theta(1/\sqrt{d})$ coordinate size, and $\gb_t$ has $\Theta(1/\sqrt{d})$ coordinate size, according to~\eqref{eq:grad_k_write}, we have
\[
    \Big(\frac{\partial\mathcal{L}}{\partial\kb_t}\Big)_l\Bigg|_{\text{write}}
    = \Theta(1)\cdot\Theta\Big(\frac{1}{\sqrt{d}}\Big) \cdot\Theta(1)
    = \Theta\Big(\frac{1}{\sqrt{d}}\Big).
\]
We now compute the erase-term contribution. The erase term is $\Eb_t = \alpha_t\Sbb_{t-1}(\Ib - \beta_t\kb_t\kb_t^{\top})$. Differentiating $(\Sbb_t)_{al} = (\Eb_t)_{al} + \beta_t(\vb_t)_a(\kb_t)_l$ with respect to $(\kb_t)_l$ gives two sub-terms:
\begin{itemize}
    \item The column of $\Sbb_{t-1}$ contracted with gradient:
\[
    \Big(\frac{\partial\mathcal{L}}{\partial\kb_t}\Big)_l\Bigg|_{\text{erase,1}}
    = -\alpha_t\beta_t \sum_a \frac{\partial\mathcal{L}}{\partial(\Sbb_t)_{al}}\sum_j (\Sbb_{t-1})_{aj}(\kb_t)_j
    = -\alpha_t\beta_t(\qb_t)_l\,\langle\gb_t, \Sbb_{t-1}\kb_t\rangle.
\]
Since $(\Sbb_{t-1}\kb_t)_i$ has variance $d\cdot\Theta(1/d)\cdot\Theta(1/d) = \Theta(1/d)$, i.e. $\Theta(1/\sqrt{d})$ per coordinate, the inner product $\langle\gb_t, \Sbb_{t-1}\kb_t\rangle$ has variance $d\cdot\Theta(1/d)\cdot\Theta(1/d) = \Theta(1/d)$, so it is $\Theta(1/\sqrt{d})$. Combined with $(\qb_t)_l = \Theta(1/\sqrt{d})$, this sub-term is $\Theta(1/d)$.

\item The diagonal of $\Sbb_{t-1}$ contracted with gradient.

The second sub-term arises from the product rule applied to $(\Sbb_{t-1}\kb_t)_a = \sum_m (\Sbb_{t-1})_{am}(\kb_t)_m$; differentiating the factor $(\kb_t)_l$ inside the sum gives
\[
    \Big(\frac{\partial\mathcal{L}}{\partial\kb_t}\Big)_l\Bigg|_{\text{erase,2}}
    = -\alpha_t\beta_t \sum_{a,c} \frac{\partial\mathcal{L}}{\partial(\Sbb_t)_{ac}}\cdot(\Sbb_{t-1})_{al}\cdot(\kb_t)_c
    = -\alpha_t\beta_t\langle \qb_t,\kb_t\rangle\,(\Sbb_{t-1}^{\top}\gb_t)_l.
\]
Since $(\Sbb_{t-1}^{\top}\gb_t)_l = \sum_a (\Sbb_{t-1})_{al}(\gb_t)_a$ has $\Theta(1/\sqrt{d})$ coordinate size, and $\langle \qb_t,\kb_t\rangle = \Theta(1/\sqrt{d})$, this sub-term is $\Theta(1/d)$ per coordinate.
\end{itemize}
Both erase sub-terms are at most $\Theta(1/d)$, smaller than the write term $\Theta(1/\sqrt{d})$. The dominant contribution therefore comes from the write term, therefore, $\left\|\frac{\partial\mathcal{L}}{\partial\kb_t}\right\|_2 = \Theta(1)$, i.e., with $\Theta\Big(\frac{1}{\sqrt{d}}\Big)$ coordinate size. Moreover, $\kb_t$ affects $\Sbb_t$ and, through the recurrent BPTT chain, all future states $\Sbb_{t''}$ for $t'' > t$. Each such future contribution has the same $d$-order as the direct term above, and Assumption~\ref{ass:short_memory} ensures their sum is $O(L_\mathrm{eff})$ times the direct term, which is an $O(1)$ factor under the short-memory assumption.

We now focus on the L2-normalization $\kb_t = \tilde{\kb}_t/\|\tilde{\kb}_t\|_2$. The Jacobian of L2-normalization is $\Jb_k = (\Ib - \kb_t\kb_t^{\top})/\|\tilde{\kb}_t\|_2$, which projects onto the hyperplane orthogonal to $\kb_t$ and scales by $1/\|\tilde{\kb}_t\|_2 = \Theta(1/\sqrt{d})$. Since $\partial\mathcal{L}/\partial\kb_t$ and $\kb_t$ are approximately independent at initialization, the projection loses negligible magnitude:
\begin{align*}
    \left\|\frac{\partial\mathcal{L}}{\partial\tilde{\kb}_t}\right\|_2
    = \Theta\Big(\frac{1}{\sqrt{d}}\Big)\cdot\Theta(1)
    = \Theta\Big(\frac{1}{\sqrt{d}}\Big),
\end{align*}
so each coordinate of $\partial\mathcal{L}/\partial\tilde{\kb}_t$ is
$\Theta(1/d)$.  The gradient of $\Wb_k$ is therefore
\[
    \frac{\partial\mathcal{L}}{\partial(\Wb_k)_{ij}} = \frac{\partial\mathcal{L}}{\partial(\tilde{\kb}_t)_i}  \cdot(\xb_t)_j = \Theta\Big(\frac{1}{d}\Big)\cdot\Theta(1) = \Theta\Big(\frac{1}{d}\Big),
\]
and an identical feature-learning analysis as for $\Wb_v$ gives
\begin{align*}
    \boxed{\eta_k = \Theta(1).}
\end{align*}

\paragraph{Gradient of the query projection $\Wb_q$.}
The gradient of $\mathcal{L}$ with respect to the normalized $\qb_t$ is
\[ \frac{\partial\mathcal{L}}{\partial(\qb_t)_j} = (\Sbb_t^{\top}\gb_t)_j = \sum_i (\Sbb_t)_{ij}(\gb_t)_i.
\]
With $(\Sbb_t)_{ij} = \Theta(1/\sqrt{d})$ and $(\gb_t)_i = \Theta(1/\sqrt{d})$, each entry of $\Sbb_t$ is approximately independent and zero-mean with variance $\Theta(1/d)$. Thus, the product $(\Sbb_t)_{ij}(g_t)_i$ has variance $\Theta(1/d^2)$. Summing over $d$ terms gives $\Var[(\Sbb_t^{\top}\gb_t)_j] = d\cdot\Theta(1/d)\cdot\Theta(1/d) = \Theta(1/d)$, i.e.\ $\partial\mathcal{L}/\partial\qb_t = \Theta(1/\sqrt{d})$ per coordinate. The L2-normalization Jacobian for $\qb_t$ has the same spectral norm $\Theta(1/\sqrt{d})$ as for $\kb_t$, therefore:
\[\frac{\partial\mathcal{L}}{\partial(\tilde{\qb}_t)_i}= \Theta\Big(\frac{1}{d}\Big),\qquad\frac{\partial\mathcal{L}}{\partial(\Wb_q)_{ij}}= \Theta\Big(\frac{1}{d}\Big).
\]
The feature-learning condition yields
\begin{align*}
    \boxed{\eta_q = \Theta(1),}
\end{align*}
confirming that all three $d\times d$ projection matrices share the same SGD learning rate scaling $\Theta(1)$.

\subsection{Gradient of the Gating}
We now show that the gating weights require a different learning rate scaling from the projection matrices, which is a distinctive feature of the SGD formulation absent in the AdamW case.

\paragraph{Gating weight $\Wb_\alpha$.}
Note that $\alpha_t$ affects not only $\Sbb_t$ but also all future states $\Sbb_{t'}$ for $t' > t$ through the recurrence chain~\eqref{eq:update_rule}. However, under Assumption~\ref{ass:short_memory} those future contributions are $O(L_\mathrm{eff}) = O(1)$ times the direct contribution and do not change the $d$-scaling of the gradient.

By the chain rule via the latent state $\mathbf{S}_t$:
\begin{align*}
    \frac{\partial \mathcal{L}}{\partial \alpha_t} = \text{Tr}\left( \left(\frac{\partial \mathcal{L}}{\partial \mathbf{S}_t}\right)^\top \frac{\partial \mathbf{S}_t}{\partial \alpha_t} \right)
\end{align*}
Since $\frac{\partial \mathcal{L}}{\partial \mathbf{S}_t} = \mathbf{g}_t \mathbf{q}_t^\top$ and $\frac{\partial \mathbf{S}_t}{\partial \alpha_t} = \mathbf{S}_{t-1}(\mathbf{I} - \beta_t \mathbf{k}_t \mathbf{k}_t^\top)$, we have
\begin{align*}
    \frac{\partial \mathcal{L}}{\partial \alpha_t} = \text{Tr}\left( \mathbf{q}_t \mathbf{g}_t^\top \mathbf{S}_{t-1}(\mathbf{I} - \beta_t \mathbf{k}_t \mathbf{k}_t^\top) \right) = \underbrace{\mathbf{g}_t^\top \mathbf{S}_{t-1} \mathbf{q}_t}_{\text{Term 1}} - \underbrace{\beta_t (\mathbf{g}_t^\top \mathbf{S}_{t-1} \mathbf{k}_t) \langle \mathbf{k}_t, \mathbf{q}_t \rangle}_{\text{Term 2}}.
\end{align*}

For Term 1, $(\mathbf{S}_{t-1} \mathbf{q}_t)_i = \sum_j (\mathbf{S}_{t-1})_{ij} (\mathbf{q}_t)_j$. Since both are zero-mean independent variables of order $\Theta(1/\sqrt{d})$, we have $\mathbf{S}_{t-1} \mathbf{q}_t = \Theta(1/\sqrt{d})$ per coordinate. Moreover, since $\mathbf{g}_t$ is $\Theta(1/\sqrt{d})$ per coordinate, finally we arrive that Term 1 is $\Theta(1/\sqrt{d})$. For Term 2, similarly we have $\mathbf{g}_t^\top \mathbf{S}_{t-1} \mathbf{k}_t = \Theta(1/\sqrt{d})$. Multiplying by $\langle \mathbf{k}_t, \mathbf{q}_t \rangle = \Theta(1/\sqrt{d})$ yields $\Theta(1/d)$. Therefore, 
\begin{align*}
    \frac{\partial \mathcal{L}}{\partial \alpha_t}=\Theta\left(\frac{1}{\sqrt{d}}\right).
\end{align*}
Since $\partial\alpha_t/\partial z_{\alpha,t} = \Theta(1)$, finally we have
\begin{align}
    \frac{\partial \mathcal{L}}{\partial z_{\alpha, t}} = \frac{\partial \mathcal{L}}{\partial \alpha_t} \cdot \frac{\partial \alpha_t}{\partial z_{\alpha, t}} = \Theta\left(\frac{1}{\sqrt{d}}\right) \cdot \Theta(1) = \Theta\left(\frac{1}{\sqrt{d}}\right)\label{eq:grad_zalpha}
\end{align}
Then the gradient per entry of $\Wb_\alpha$ is
\[
    \frac{\partial\mathcal{L}}{\partial(\Wb_\alpha)_j}
    = \frac{\partial\mathcal{L}}{\partial z_{\alpha,t}}\cdot(\xb_t)_j
    = \Theta\Big(\frac1{\sqrt{d}}\Big)\cdot\Theta(1)
    = \Theta\Big(\frac1{\sqrt{d}}\Big).
\]
To enable feature learning, the scalar pre-activation $z_{\alpha}$ must move by $\Theta(1)$ after one gradient step. The change in $z_{\alpha}$ is
\begin{align*}
    \Delta z_{\alpha} = \Delta \mathbf{W}_\alpha \mathbf{x}_t = -\eta_\alpha \sum_{j=1}^d \frac{\partial \mathcal{L}}{\partial (\mathbf{W}_\alpha)_j} (\mathbf{x}_t)_j = -\eta_\alpha \frac{\partial \mathcal{L}}{\partial z_{\alpha, t}} \sum_{j=1}^d (\mathbf{x}_t)_j^2.
\end{align*}
Because $\sum_j (\mathbf{x}_t)_j^2 = \|\mathbf{x}_t\|_2^2 = \Theta(d)$, we have
\begin{align*}
    \Delta z_{\alpha} = \eta_\alpha \cdot \Theta\left(\frac{1}{\sqrt{d}}\right) \cdot \Theta(d) = \eta_\alpha \Theta(\sqrt{d}).
\end{align*}

For $\Delta z_{\alpha} = \Theta(1)$, the learning rate must be set to
\begin{align*}
\boxed{\eta_\alpha = \Theta\Big(\frac{1}{\sqrt{d}}\Big)= \Theta\Big(\frac{1}{\sqrt{n_{\ell-1}}}\Big).}
\end{align*}

\paragraph{Gating weight $\Wb_\beta$.}
Similarly, $\beta_t$ affects all future states through the recurrence, but under Assumption~\ref{ass:short_memory} we need only the direct contribution to $\Sbb_t$. The gradient of $\beta_t$ combines contributions from both the write term and the erase term of~\eqref{eq:update_rule}:
\[
    \frac{\partial\mathcal{L}}{\partial\beta_t}
    = \sum_{a,b}\frac{\partial\mathcal{L}}{\partial(\Sbb_t)_{ab}}
      \cdot\frac{\partial(\Sbb_t)_{ab}}{\partial\beta_t}
    = \underbrace{\langle \mathbf{g}_t, \mathbf{v}_t \rangle \langle \mathbf{q}_t,\mathbf{k}_t\rangle}_{\text{Write Term}} - \underbrace{\alpha_t \langle\gb_t,\Sbb_{t-1}\kb_t\rangle\langle\qb_t,\kb_t\rangle }_{\text{Erase Term}}.
\]
Since $\langle\gb_t,\vb_t\rangle\langle\qb_t,\kb_t\rangle = \Theta(1)\cdot\Theta(1/\sqrt{d}) = \Theta(1/\sqrt{d})$, the write contribution is $\partial(\Sbb_t)_{ab}/\partial\beta_t\big|_{\text{write}} = (\vb_t)_a(\kb_t)_b$. And since $\langle\gb_t,\Sbb_{t-1}\kb_t\rangle\langle\qb_t,\kb_t\rangle = \Theta(1/\sqrt{d})\cdot\Theta(1/\sqrt{d}) = \Theta(1/d)$, the erase term is $\partial(\Sbb_t)_{ab}/\partial\beta_t\big|_{\text{erase}} = -\alpha_t\sum_c(\Sbb_{t-1})_{ac}(\kb_t)_c(\kb_t)_b=-\alpha_t (\Sbb_{t-1}\kb_t)_a(\kb_t)_b$. The write term therefore dominates and
\begin{align*}
    \frac{\partial\mathcal{L}}{\partial\beta_t}
    = \Theta\Big(\frac{1}{\sqrt{d}}\Big).
\end{align*}

For $z_{\beta,t} := \Wb_\beta\xb_t$, since $\frac{\partial\beta_t}{\partial z_{\beta,t}}=\Theta(1)$, then the gradient of the pre-activation is
\begin{align*}
    \frac{\partial\mathcal{L}}{\partial z_{\beta,t}}
    = \frac{\partial\beta_t}{\partial z_{\beta,t}}\cdot\frac{\partial\mathcal{L}}{\partial\beta_t}
    = \Theta(1)\cdot\Theta\Big(\frac{1}{\sqrt{d}}\Big)
    = \Theta\Big(\frac{1}{\sqrt{d}}\Big).
\end{align*}
Finally, the gradient per entry of $\Wb_\beta$ is
\[
    \frac{\partial\mathcal{L}}{\partial(\Wb_\beta)_j}
    = \frac{\partial\mathcal{L}}{\partial z_{\beta,t}}\cdot(\xb_t)_j
    = \Theta\Big(\frac{1}{\sqrt{d}}\Big)\cdot\Theta(1)
    = \Theta\Big(\frac{1}{\sqrt{d}}\Big),
\]
and the same feature-learning analysis as $\Wb_\alpha$ gives
\begin{align*}
    \boxed{\eta_\beta = \Theta\Big(\frac{1}{\sqrt{d}}\Big)
    = \Theta\Big(\frac{1}{\sqrt{n_{\ell-1}}}\Big).}
\end{align*}

\paragraph{Scalar parameters.}
Both $a_{\log}$ and $b$ enter through the gating pre-activations as additive scalars. Using the exact chain rule on $\alpha_t = e^{g_t}$,
\[\frac{\partial\alpha_t}{\partial a_{\log}} = \frac{\partial e^{g_t}}{\partial a_{\log}}=e^{g_t}\frac{\partial g_t}{\partial a_{\log}}=\alpha_tg_t=\Theta(1),\]
so $\partial\mathcal{L}/\partial a_{\log} = (\partial\mathcal{L}/\partial\alpha_t)\cdot\Theta(1)$. From the argument in~\eqref{eq:grad_zalpha}, $\partial\mathcal{L}/\partial\alpha_t = \Theta(1/\sqrt{d})$. Hence $\partial\mathcal{L}/\partial a_{\log} = \Theta(1/\sqrt{d})$, and an SGD step changes $a_{\log}$ by $\eta_{\mathrm{scal}}\cdot\Theta(1/\sqrt{d})$. Since $a_{\log}$ enters $\alpha_t$ multiplicatively, a $\Theta(1)$ change in $\alpha_t$ via $a_{\log}$ requires $\eta_{\mathrm{scal}} = \Theta(\sqrt{d})$.

\subsection{Summary}
According to the $\mu$P assumptions, the learning rate for other parameters can be set in the same way as Table 8 in~\citet{yang2022tensor}. In summary, the complete $\mu$P formulation for Gated Delta Net with SGD is summarized in Table~\ref{tab:gdn_sgd_formulation}.

\begin{table*}[htbp]
\caption{$\mu$P formulation of Gated Delta Net under SGD. Weights have shape $\RR^{n_\ell\times n_{\ell-1}}$; for input weights and biases, $n_{\ell-1}=\Theta(1)$. All initialization variances and forward multipliers are identical to the AdamW table (Table~\ref{tab:gdn_formulation}), and only the learning-rate row differs. The {\color{gray}gray} factors are the differences between original scaling law in~\citet{yang2022tensor} and our proposed formulation.}
\label{tab:gdn_sgd_formulation}

\begin{center}
\begin{small}
\resizebox{\linewidth}{!}{%
\begin{tabular}{lccccc}
\toprule
& \makecell{Input weights \& \\ all other biases} & Output weights & \makecell{Hidden weights \\(except $\Wb_\alpha,\Wb_\beta$)} & $\Wb_\alpha,\Wb_\beta$ & $a_{\log}$ and $b$ \\
\midrule
Initial variance & $\frac{1}{n_{\ell-1}}$ & $1$ & $\frac{1}{n_{\ell-1}}$ & $\frac{1}{n_{\ell-1}}$ & $1$ \\[6pt]
Multiplier & $1$ & $\frac{1}{n_{\ell-1}}$ & $1$ & $1$ & $1$ \\[6pt]
SGD LR & $n_{\ell}$ & $n_{\ell-1}$ & $1$ & $\frac{1}{\sqrt{n_{\ell-1}}}$ {\color{gray}(1)} & $\sqrt{n_{\ell-1}}$ {\color{gray}(1)} \\
\bottomrule
\end{tabular}}
\end{small}
\end{center}
\end{table*}
\section{Experiments}
\label{sec:exp}
\subsection{Experiment details}
We implement LLM pre-training experiments to validate our $\mu$P derivation. All models use 8 layers and 6 attention heads. We test five model widths $d \in \{256, 512, 1024, 1536\}$ for AdamW~\citep{loshchilov2017decoupled} and $d \in \{256, 512, 768, 1024\}$ for SGD optimizer, which correspond to parameter counts ranging from approximately 21M to 342M (non-embedding). 

\paragraph{Architectural parameters.} We refer to~\citep{yang2025gated} and its official repository for the implementation of GDN and re-implement it on \texttt{nanoGPT} training framework~\citep{Karpathy2022}. In detail, we set the head dimension of queries and keys to $d/8$ and that of values to $d/4$. And we set the kernel size of short convolutions in queries and keys to 4. Additionally, the intermediate size of MLP is set to $4d$, and we tied the input and output embeddings. 

\paragraph{Initialization and optimizer.} For the base model with $d_0=256$, the embedding layer and all input projections are initialized with standard deviation 0.02, which also applies to larger models for SP. In contrast, we initialize large models under original $\mu$P and our proposed $\mu$P according to Tables~\ref{tab:gdn_formulation} and~\ref{tab:gdn_sgd_formulation}. The scalar gating parameters $a_{\log}$ and $b$ follow the scheme of~\citet{yang2025gated}: $a_{\log}\sim\mathrm{Uniform}(0,16)$ and $b$ is set as described in Section~\ref{sec:mup_analysis}. We apply a gradient clipping to 1.0 and set Dropout ratio~\citep{srivastava2014dropout} to 0.0. The minimum learning rate is fixed to 5e-5 throughout. All runs use a cosine learning-rate schedule with 2,000 warmup steps.

For AdamW experiments, we use a weight decay of $0.1$ and set $(\beta_1,\beta_2)=(0.9,0.95)$. And for SGD experiments, we use SGD with Nesterov momentum~\citep{nesterov1983method} with a momentum of 0.98, since we notice there is great instability when using original SGD optimizer. For both optimizers, we use the same learning rates for all the modules in models with $d_0=256$ and all models under SP, and applies different learning rates according to the scaling laws in Tables~\ref{tab:gdn_formulation} and~\ref{tab:gdn_sgd_formulation} in $\mu$P experiments.

We train models with each width at 5-7 different learning rates log-spaced with increased density near optimal learning rates. The learning rate search grid ranges between 1e-3 and 2e-2 for AdamW and between 1e-1 and 1 for SGD experiments. And we fix the training seed to 42.

\paragraph{Data and compute.} We train on the FineWeb-Edu 100B dataset~\citep{lozhkov2024fineweb-edu} for 20k steps with a global batch size of 480 sequences and a sequence length of 1024 (approximately 9.83B tokens in total). Moreover, we use 1 NVIDIA H100 80GB HBM3 GPU for all the experiments.

\subsection{Experiment results}
\begin{figure*}[h]
    \centering
    \begin{subfigure}[b]{0.45\linewidth}
        \centering
        \includegraphics[width=\linewidth]{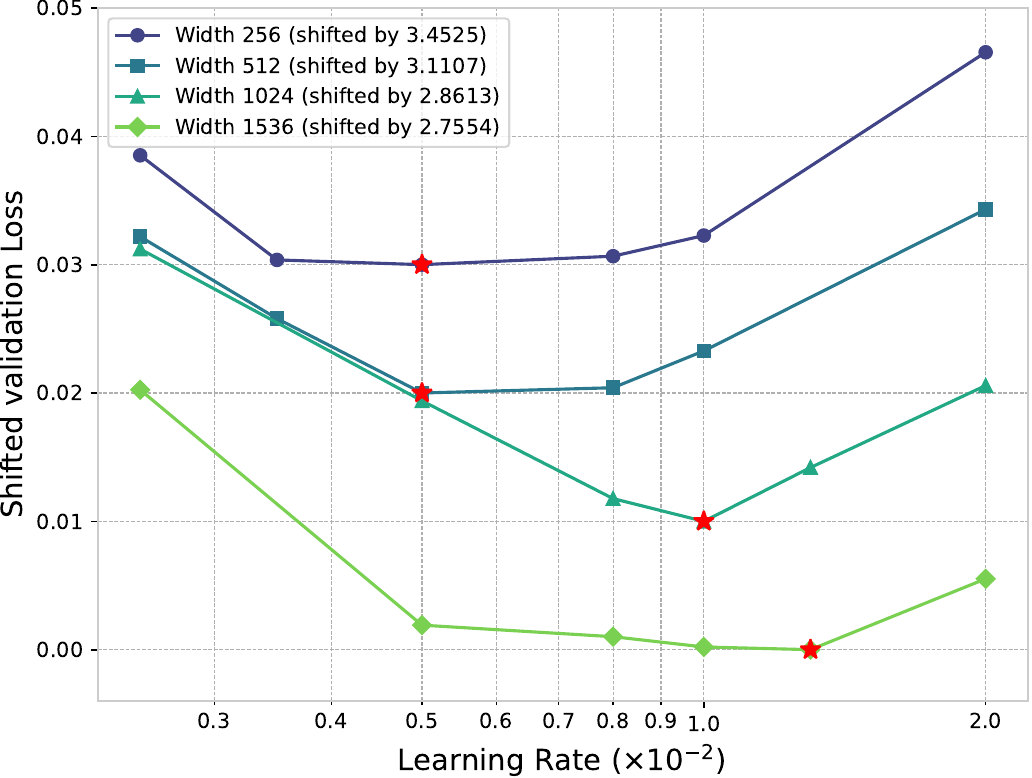}
        \caption{Standard Parametrization (SP)}
    \end{subfigure}
    \begin{subfigure}[b]{0.45\linewidth}
        \centering
        \includegraphics[width=\linewidth]{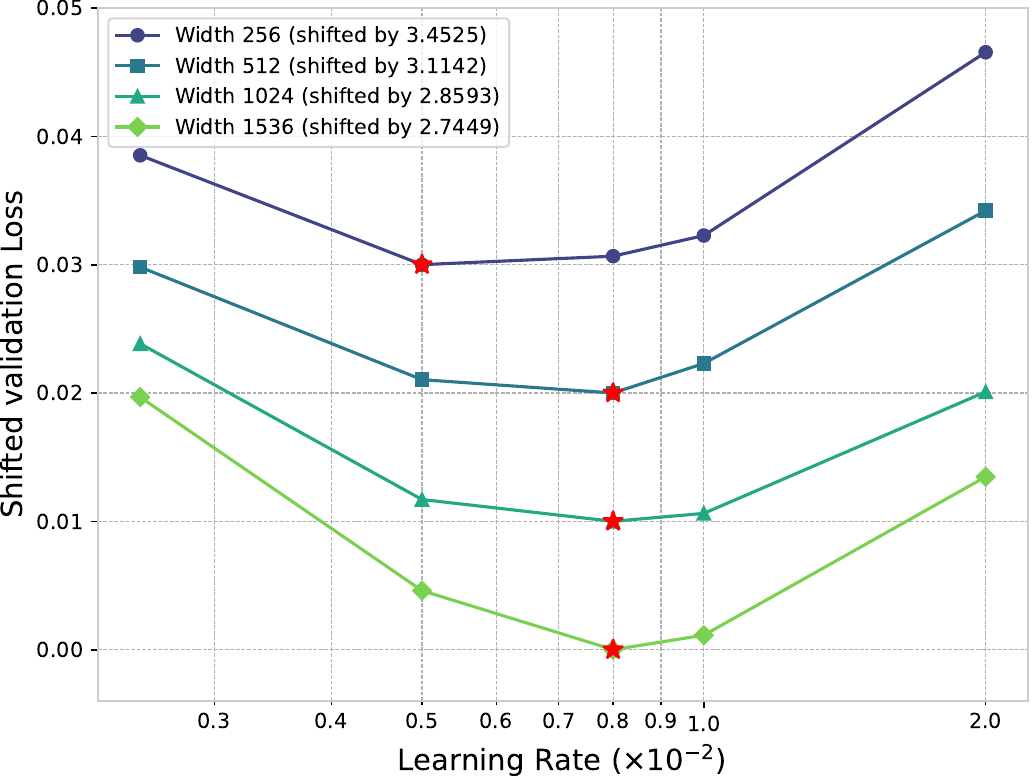}
        \caption{$\mu$P configuration}
    \end{subfigure}
    \caption{Shifted validation loss for Gated Delta Network trained with AdamW under varying peak learning rates and model widths.}
    \label{fig:gdn}
\end{figure*}

\begin{figure*}[h]
    \centering
    \begin{subfigure}[b]{0.32\linewidth}
        \centering
        \includegraphics[width=\linewidth]{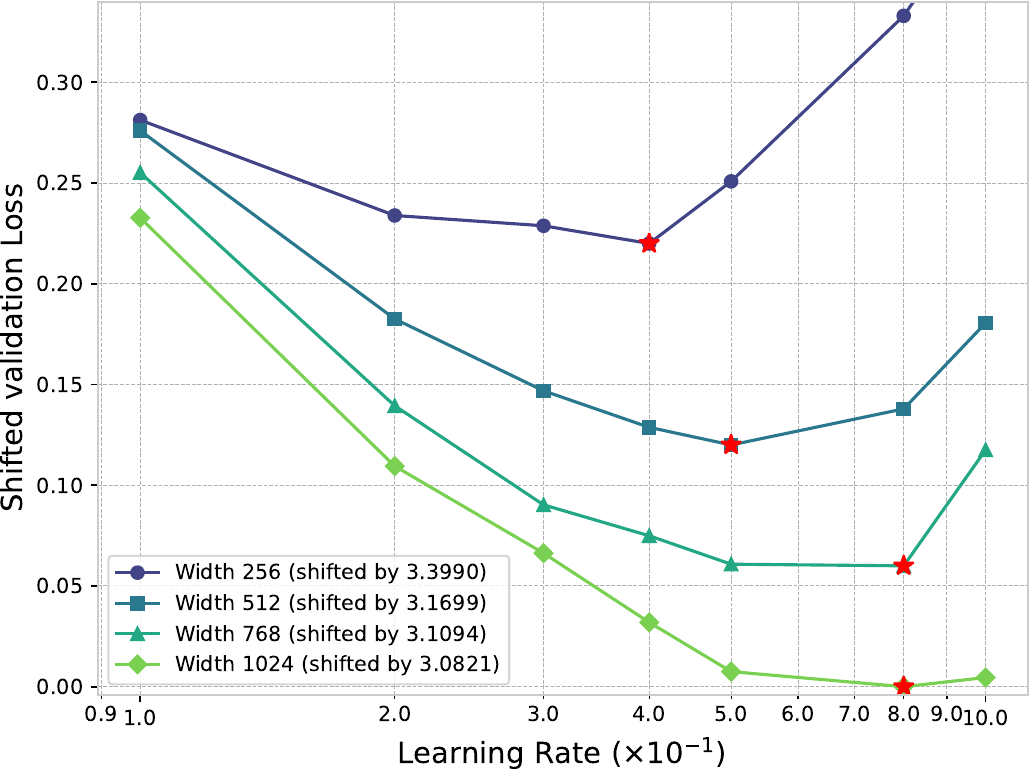}
        \caption{Standard Parametrization (SP)}
    \end{subfigure}
    \hfill
    \begin{subfigure}[b]{0.32\linewidth}
        \centering
        \includegraphics[width=\linewidth]{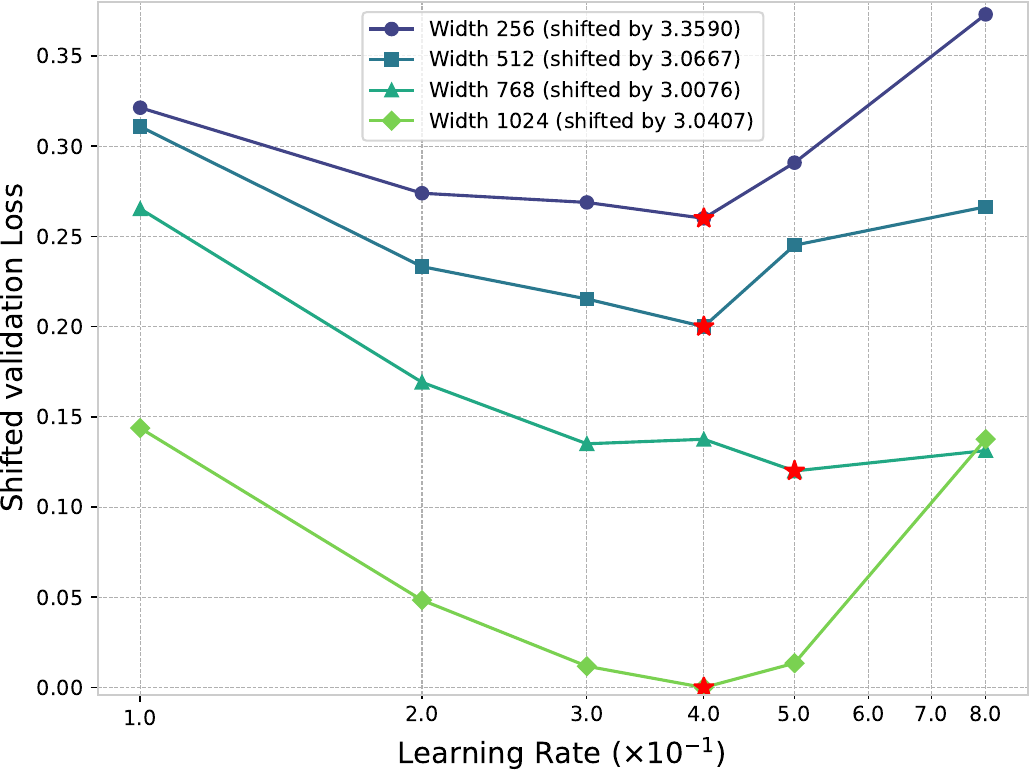}
        \caption{Original $\mu$P configuration}
    \end{subfigure}
    \hfill
    \begin{subfigure}[b]{0.32\linewidth}
        \centering
        \includegraphics[width=\linewidth]{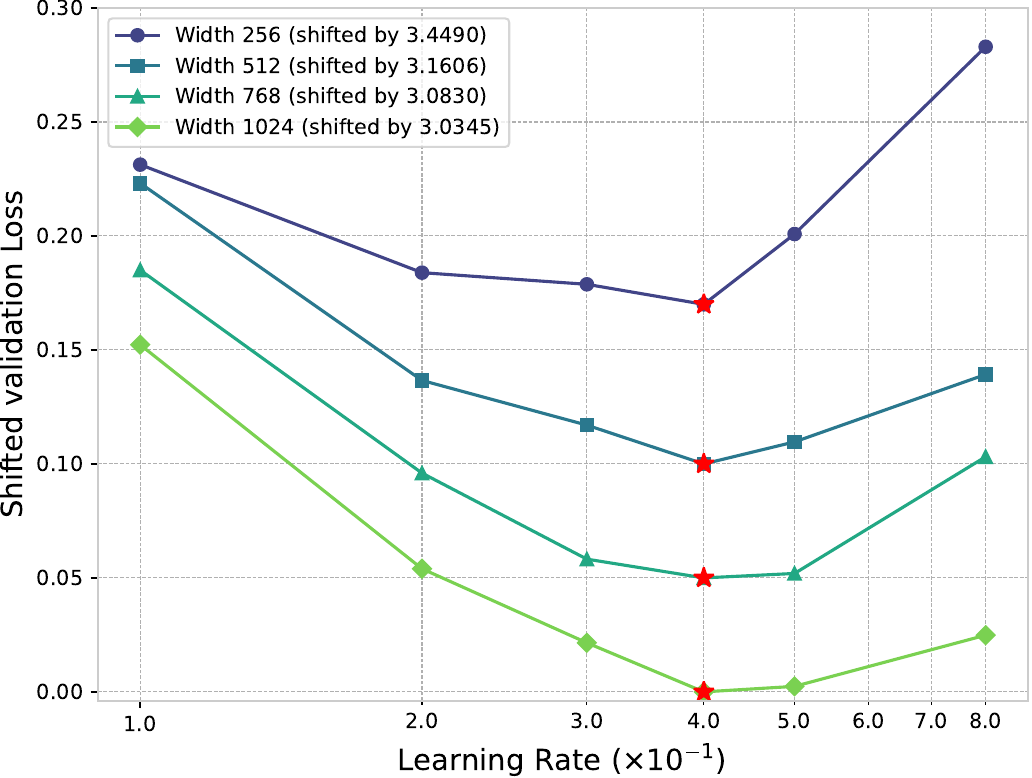}
        \caption{Our $\mu$P configuration}
    \end{subfigure}
    \caption{Shifted validation loss (loss minus the best-achieved loss at width $d=1024$) for Gated Delta Network trained with SGD under varying peak learning rates and model widths.}
    \label{fig:gdn_sp}
\end{figure*}

The final validation losses for models with different widths and peak learning rates under the $\mu$P and SP configurations are shown in Figures~\ref{fig:gdn} and~\ref{fig:gdn_sp} for AdamW and SGD, respectively. To remove the trivial width-dependence of the absolute loss, we report \emph{shifted} validation loss, defined as the difference from the optimal loss value among all the experiments on the models with the same width but different learning rates. 

For AdamW, the optimal learning rate is consistently the same across all 4 model widths under $\mu$P, demonstrating zero-shot learning-rate transfer. While under SP, the optimal learning rate shifts substantially with width, confirming that SP fails to support feature learning at scale. SGD experiments show the same qualitative pattern. Under SP, the optimal learning rate does not transfer across widths and under original $\mu$P configuration, it varies a lot. And in our $\mu$P configuration, the optimal learning rate transfers perfectly. These results validate that our theory works well in practice.

\section{Conclusion}

We have derived the $\mu$P-style parametrization for Gated Delta Networks. Our analysis reveals that under SGD, scalings of the gating weight matrices and the scalar gating parameters are different from the standard $\mu$P law. LLM pre-training experiments confirm that our $\mu$P formulation achieves zero-shot learning-rate transfer under both AdamW and SGD, while standard parametrization fails to transfer, empirically validating the correctness of our theoretical derivation. And we hope our derivations would enlighten further research in the scaling laws of other linear or hybrid architectures.

\section*{Code Availability}
The code for this paper can be accessed in \url{https://github.com/lauyikfung/gated_delta_net_mup}.

\section*{Acknowledgement}
Thank Fetch Compute program for their support of compute resources. Thank Songlin Yang for discussion. Thank Amazon Trainium scholarship project for their funding support. And the code specific for Trainium chips can be accessed in \url{https://github.com/lauyikfung/Amazon_Trainium_Optimizer/tree/main/gdn_mup_code}.

\bibliography{reference}
\bibliographystyle{plainnat}

\clearpage
\renewcommand{\appendixpagename}{\centering \huge Appendix}
\appendixpage
\counterwithin{theorem}{section}

\startcontents[section]
\printcontents[section]{l}{1}{\setcounter{tocdepth}{2}}
\clearpage
\appendix
\section{Additional derivations in the backward process for SGD}
\label{sec:additional_backward_derivation}
In this section, we provide the detailed analysis of BPTT term in this section.
\subsection{Derivation of the cumulative latent space}\label{sec:cumulative_latent_space}
In Section~\ref{coordinate_size_of_latent_state}, we uses the argument of~\citet{vankadara2024feature} to direct achieve that $\Sbb_t$ has $\Theta(1/\sqrt{d})$ coordinate size. Here we show the detailed derivations of this conclusion. The state update rule is:
\begin{align*}
    \mathbf{S}_t = \alpha_t \mathbf{S}_{t-1} - \alpha_t \beta_t \mathbf{S}_{t-1} \mathbf{k}_t \mathbf{k}_t^\top + \mathbf{U}_t,
\end{align*}
where the write matrix is $\mathbf{U}_t = \beta_t \mathbf{v}_t \mathbf{k}_t^\top$. From the $\mu$P initialization, we know that $(\mathbf{v}_t)_i = \Theta(1)$, $(\mathbf{k}_t)_j = \Theta(1/\sqrt{d})$ and $\alpha_t, \beta_t \in (0, 1)$ are bounded scalars, which can be treated as $\Theta(1)$ independent variables. And both $(\mathbf{v}_t)_i$ and $(\mathbf{k}_t)_j$ are with zero mean values. Therefore,
\begin{align*}
    \mathbb{E}[(\mathbf{U}_t)_{ij}^2] = \mathbb{E}[\beta_t^2 (\mathbf{v}_t)_i^2 (\mathbf{k}_t)_j^2] = \Theta(1) \cdot \Theta(1) \cdot \Theta\left(\frac{1}{d}\right) = \Theta\left(\frac{1}{d}\right).
\end{align*}
Therefore, $\mathbf{U}_t$ has a coordinate size of $\Theta(1/\sqrt{d})$.

Let $V_t = \mathbb{E}[(\mathbf{S}_t)_{ij}^2]$ be the element-wise variance of the state matrix. As a standard mean-field assumption at initialization, we assume $\mathbf{S}_{t-1}$ is statistically independent of the current inputs $\mathbf{v}_t$ and $\mathbf{k}_t$. Then,
\begin{align*}
    (\mathbf{S}_t)_{ij} = \alpha_t (\mathbf{S}_{t-1})_{ij} - \alpha_t \beta_t \sum_l (\mathbf{S}_{t-1})_{il} (\mathbf{k}_t)_l (\mathbf{k}_t)_j + (\mathbf{U}_t)_{ij}.
\end{align*}

Since $\mathbf{U}_t$ depends on $\mathbf{v}_t$, which has zero mean and is independent of $\mathbf{S}_{t-1}$ and $\mathbf{k}_t$, the cross-terms between $\mathbf{U}_t$ and the other terms vanish when we take the expected square. Therefore,
\begin{align*}
    V_t = \mathbb{E} \left[ \left( \alpha_t (\mathbf{S}_{t-1})_{ij} - \alpha_t \beta_t \sum_l (\mathbf{S}_{t-1})_{il} (\mathbf{k}_t)_l (\mathbf{k}_t)_j \right)^2 \right] + \mathbb{E}[(\mathbf{U}_t)_{ij}^2].
\end{align*}
For the squared expectation term, we have
\begin{align*}
    \mathbb{E}[\alpha_t^2 (\mathbf{S}_{t-1})_{ij}^2] = \mathbb{E}[\alpha_t^2] V_{t-1},
\end{align*}
and
\begin{align*}
    -2 \mathbb{E} \left[ \alpha_t^2 \beta_t (\mathbf{S}_{t-1})_{ij} \sum_l (\mathbf{S}_{t-1})_{il} (\mathbf{k}_t)_l (\mathbf{k}_t)_j \right]= -2 \mathbb{E}[\alpha_t^2 \beta_t] \mathbb{E}[(\mathbf{S}_{t-1})_{ij}^2] \frac{1}{d} = -2 \mathbb{E}[\alpha_t^2 \beta_t] V_{t-1} \frac{1}{d},
\end{align*}
where the second equality holds since $\mathbf{k}_t$ has independent zero-mean entries, and then the expectation over $\mathbf{k}_t$ is zero unless $l = j$. And for the remaining term, we have
\begin{align*}
    \mathbb{E} \left[ \alpha_t^2 \beta_t^2 \left( \sum_l (\mathbf{S}_{t-1})_{il} (\mathbf{k}_t)_l \right)^2 (\mathbf{k}_t)_j^2 \right]&=\mathbb{E} \left[ \alpha_t^2 \beta_t^2 \big( \sum_{l, m} (\mathbf{S}_{t-1})_{il} (\mathbf{S}_{t-1})_{im} (\mathbf{k}_t)_l (\mathbf{k}_t)_m \big) (\mathbf{k}_t)_j^2 \right]\\
    &=\mathbb{E} [ \alpha_t^2 \beta_t^2 (\sum_l (\mathbf{S}_{t-1})_{il}^2 (\mathbf{k}_t)_l^2) (\mathbf{k}_t)_j^2 ]\\
    &= \mathbb{E}[\alpha_t^2 \beta_t^2] \left( d \cdot V_{t-1} \cdot \frac{1}{d^2} \right) \\
    &= \mathbb{E}[\alpha_t^2 \beta_t^2] V_{t-1} \frac{1}{d},
\end{align*}
where the second equality holds by the independence within the entries in $\kb_t$. Finally we have
\begin{align*}
    V_t &= \mathbb{E}[\alpha_t^2] V_{t-1} - \frac{2 \mathbb{E}[\alpha_t^2 \beta_t] - \mathbb{E}[\alpha_t^2 \beta_t^2]}{d} V_{t-1} + \Theta\left(\frac{1}{d}\right)\\
    &=V_{t-1} \cdot \underbrace{\mathbb{E} \left[ \alpha_t^2 \left( 1 - \frac{2\beta_t - \beta_t^2}{d} \right) \right]}_{\gamma} + \Theta\left(\frac{1}{d}\right).
\end{align*}
Assume $\mathbb{E}[\alpha_t^2] = c^2 < 1$. 
Because $d$ is relatively large, the $(2\beta_t - \beta_t^2)/d$ term is small. Therefore, the contraction factor $\gamma$ is strictly dictated by $\alpha_t$, and we have $\gamma \approx \mathbb{E}[\alpha_t^2] < 1$. Then the variance $V_t$ is a simple converging geometric progression:
\begin{align*}
    V_t = \gamma V_{t-1} + \Theta\left(\frac{1}{d}\right).
\end{align*}
As $t \to \infty$, it converges to the infinite sum:
\begin{align*}
    V_\infty = \frac{\Theta(1/d)}{1 - \gamma},
\end{align*}
which is $\Theta(1/d)$ since $\gamma$ is $\Theta(1)$ and strictly less than $1$. Therefore, the element-wise variance of the latent state $\mathbf{S}_t$ is $\Theta(1/d)$, and the coordinate size is $\Theta(1/\sqrt{d})$.

\section{Compatibility with AdamW}
\begin{table*}[h]
\caption{$\mu$P formulation of Gated Delta Net under AdamW optimization. Weights have shape $\RR^{n_\ell \times n_{\ell-1}}$; for input weights and biases, $n_{\ell-1} = \Theta(1)$. Under AdamW, Adam's coordinate-wise gradient normalization equalizes the effective update magnitude across all weight classes, so the gating weight matrices $\Wb_\alpha$ and $\Wb_\beta$ are subsumed into the ``Hidden weights'' column and require no special treatment.}
\label{tab:gdn_formulation}
\begin{center}
\begin{small}
\resizebox{\linewidth}{!}{%
\begin{tabular}{lcccc}
\toprule
 & Input weights \& all biases & Output weights & Hidden weights & $a_{\log}$ and $b$ \\
\midrule
Initial variance  & $\frac{1}{n_{\ell-1}}$ & $1$  & $\frac{1}{n_{\ell-1}}$ & $1$ \\
Multiplier & $1$ & $\frac{1}{n_{\ell-1}}$ & $1$ & $1$ \\
Learning rate  & $1$ & $1$  & $\frac{1}{n_{\ell-1}}$ & $1$ \\
\bottomrule
\end{tabular}}
\end{small}
\end{center}
\end{table*}
\label{sec:compat_adamw}
Under the AdamW~\citep{loshchilov2017decoupled} optimizer, the parameter update at step $t$ is proportional to the coordinate-wise exponential moving average of the gradient, so the effective coordinate-wise update magnitude is $\Theta(1)$ regardless of the gradient scale. Hence the $\mu$P learning-rate scaling is similar to that for standard architectures without modification. The complete $\mu$P formulation for Gated Delta Net trained with AdamW is summarized in Table~\ref{tab:gdn_formulation}. To rigorously validate the intuition, we will derive the scaling law of GDN when optimized by Adam(W) in the following.

\subsection{Adam(W) in the Scale-Invariant Regime}
Consider the pre-activation of the gating mechanism, $z_{\alpha, t} = \mathbf{W}_\alpha \mathbf{x}_t + b$. From our first-order analysis, the gradient of the loss with respect to this pre-activation is $\delta_{\alpha, t} := \frac{\partial \mathcal{L}}{\partial z_{\alpha, t}} = \Theta(1/\sqrt{d})$.

For any parameter $\theta$ with gradient $g = \partial \mathcal{L} / \partial \theta$, the Adam(W) update rule normalizes the gradient by its root-mean-square $v$. Since the gradients in our model are order $\Theta(1/\sqrt{d})$, which strictly dominates the standard Adam(W) $\epsilon$ parameter (e.g., $10^{-8}$) for practically sized $d$, Adam(W) operates in a scale-invariant regime. The update effectively reduces to a coordinate-wise sign descent:
\begin{equation}
    \Delta \theta_j = -\eta_\theta \frac{g_j}{\sqrt{v_j} + \epsilon} \approx -\eta_\theta \text{sign}(g_j),
    \label{eq:adam_sign_descent}
\end{equation}
where $\eta_\theta$ is the learning rate assigned to parameter $\theta$.

\subsection{Derivation for Main Projection Weights ($\mathbf{W}_q, \mathbf{W}_k, \mathbf{W}_v, \mathbf{W}_o$)}
Let $\mathbf{W} \in \mathbb{R}^{d \times d}$ denote any of the main linear projection matrices, yielding a pre-activation $\mathbf{z} = \mathbf{W} \mathbf{x}_t \in \mathbb{R}^d$. Let $\delta_i = \frac{\partial \mathcal{L}}{\partial z_i}$ be the gradient arriving at the $i$-th coordinate of the output. The gradient with respect to the weight coordinate $W_{ij}$ is:
\begin{align*}
    g_{ij} = \frac{\partial \mathcal{L}}{\partial W_{ij}} = \delta_i (\mathbf{x}_t)_j.
\end{align*}
Applying the Adam update rule \eqref{eq:adam_sign_descent}, the parameter change is:
\begin{align*}
    \Delta W_{ij} \approx -\eta_W \text{sign}\big(\delta_i (\mathbf{x}_t)_j\big) = -\eta_W \text{sign}(\delta_i) \text{sign}\big((\mathbf{x}_t)_j\big).
\end{align*}
We now compute the exact shift $\Delta z_i$ in the $i$-th coordinate of the feature caused by this update:
\begin{align}
    \Delta z_i &= \sum_{j=1}^d \Delta W_{ij} (\mathbf{x}_t)_j \nonumber \\
    &= \sum_{j=1}^d \left( -\eta_W \text{sign}(\delta_i) \text{sign}\big((\mathbf{x}_t)_j\big) \right) (\mathbf{x}_t)_j \nonumber \\
    &= -\eta_W \text{sign}(\delta_i) \sum_{j=1}^d \big|(\mathbf{x}_t)_j\big|. \label{eq:l1_norm_sum_main}
\end{align}
Under the $\mu$P initialization, each coordinate of the hidden state $\mathbf{x}_t$ possesses a magnitude of $\Theta(1)$. Consequently, the sum of absolute values in \eqref{eq:l1_norm_sum_main} (i.e., the $\ell_1$ norm of $\mathbf{x}_t$) strictly scales as $\Theta(d)$. Thus, the magnitude of the feature shift is:
\begin{align*}
    |\Delta z_i| = \eta_W \cdot \Theta(d).
\end{align*}
To enforce the feature-learning requirement $|\Delta z_i| = \Theta(1)$, the learning rate for the main projection matrices must be scaled as:
\begin{align*}
    \eta_W = \Theta\left(\frac{1}{d}\right).
\end{align*}

\subsection{Derivation for Gating Weights ($\mathbf{W}_\alpha, \mathbf{W}_\beta$)}
For the gating weight matrix $\mathbf{W}_\alpha \in \mathbb{R}^{1 \times d}$ (the derivation for $\mathbf{W}_\beta$ is identical), the gradient with respect to its $j$-th coordinate is:
\begin{align*}
    g_{\alpha, j} = \frac{\partial \mathcal{L}}{\partial (\mathbf{W}_\alpha)_j} = \delta_{\alpha, t} (\mathbf{x}_t)_j.
\end{align*}
Substituting this into the Adam update rule \eqref{eq:adam_sign_descent}, we obtain the exact parameter update:
\begin{align*}
    \Delta (\mathbf{W}_\alpha)_j = -\eta_\alpha \text{sign}\big(\delta_{\alpha, t} (\mathbf{x}_t)_j\big) = -\eta_\alpha \text{sign}(\delta_{\alpha, t}) \text{sign}\big((\mathbf{x}_t)_j\big).
\end{align*}
To satisfy the fundamental $\mu$P feature-learning condition, a single optimization step must induce a $\Theta(1)$ shift in the pre-activation $z_{\alpha, t}$. We compute this exact shift $\Delta z_{\alpha, t}$ caused by $\Delta \mathbf{W}_\alpha$:
\begin{align}
    \Delta z_{\alpha, t} &= \sum_{j=1}^d \Delta (\mathbf{W}_\alpha)_j (\mathbf{x}_t)_j \nonumber \\
    &= \sum_{j=1}^d \left( -\eta_\alpha \text{sign}(\delta_{\alpha, t}) \text{sign}\big((\mathbf{x}_t)_j\big) \right) (\mathbf{x}_t)_j \nonumber \\
    &= -\eta_\alpha \text{sign}(\delta_{\alpha, t}) \sum_{j=1}^d \big|(\mathbf{x}_t)_j\big|. \label{eq:l1_norm_sum}
\end{align}
Under the $\mu$P assumptions, each coordinate of the hidden state $\mathbf{x}_t$ possesses a magnitude of $\Theta(1)$. Consequently, the sum of absolute values in \eqref{eq:l1_norm_sum} (i.e., the $\ell_1$ norm of $\mathbf{x}_t$) strictly scales as $\Theta(d)$. Thus, the magnitude of the feature shift is:
\begin{align*}
    |\Delta z_{\alpha, t}| = \eta_\alpha \cdot \Theta(d).
\end{align*}
To enforce the feature-learning requirement $|\Delta z_{\alpha, t}| = \Theta(1)$, the learning rate must be precisely scaled as:
\begin{align*}
    \eta_\alpha = \Theta\left(\frac{1}{d}\right).
\end{align*}
This demonstrates a core property of AdamW under $\mu$P: regardless of the output dimension of the projection, the learning rate scaling is strictly dictated by the fan-in dimension ($d$). Therefore, both matrix types can be summarized as one category with a $1/n_{\ell-1}$ learning rate multiplier, perfectly aligning with Table~\ref{tab:gdn_formulation}.

\subsection{Derivation for Scalar Parameters ($a_{\log}$ and $b$)}
We now analyze the scalar bias $b$ (and equivalently, the additive scalar $a_{\log}$). The gradient with respect to $b$ is trivially the pre-activation gradient:
\begin{align*}
    g_b = \frac{\partial \mathcal{L}}{\partial b} = \delta_{\alpha, t} = \Theta\left(\frac{1}{\sqrt{d}}\right).
\end{align*}
Applying the scale-invariant Adam update, the parameter change is:
\begin{align*}
    \Delta b \approx -\eta_b \text{sign}(\delta_{\alpha, t}).
\end{align*}
Because $b$ acts as a direct additive scalar to the pre-activation ($z_{\text{new}} = \mathbf{W}_\alpha \mathbf{x}_t + b_{\text{new}}$), the resulting shift in the feature is exactly the parameter update itself:
\begin{align*}
    |\Delta z_{\alpha, t,\text{(from bias)}}| = |\Delta b| = \eta_b \cdot \Theta(1).
\end{align*}
To achieve the requisite $\Theta(1)$ feature drift, the learning rate for the scalar must be independent of the width:
\begin{align*}
    \eta_b = \Theta(1).
\end{align*}
\begin{figure*}[h]
    \centering
    \begin{subfigure}[b]{0.32\linewidth}
        \centering
        \includegraphics[width=\linewidth]{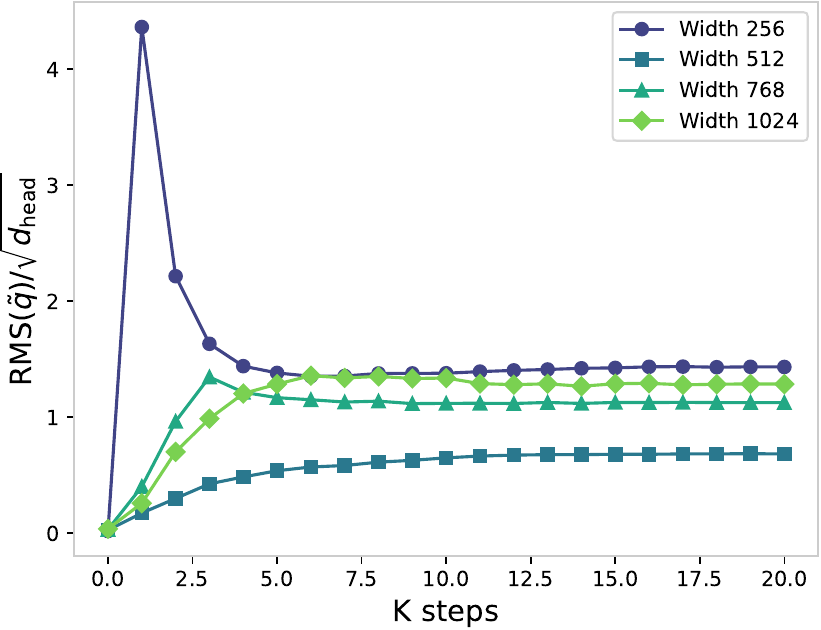}
        \caption{Standard Parametrization (SP)}
    \end{subfigure}
    \hfill
    \begin{subfigure}[b]{0.32\linewidth}
        \centering
        \includegraphics[width=\linewidth]{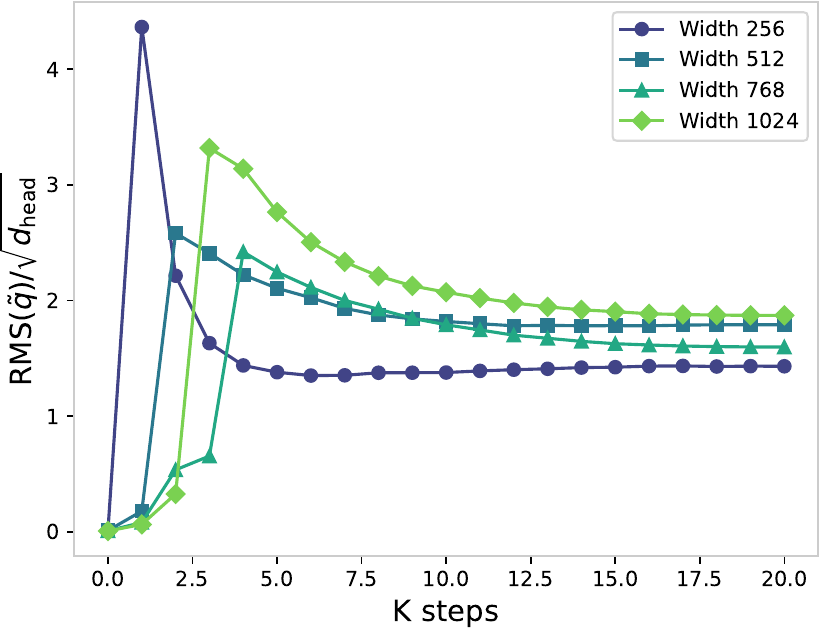}
        \caption{Original $\mu$P configuration}
    \end{subfigure}
    \hfill
    \begin{subfigure}[b]{0.32\linewidth}
        \centering
        \includegraphics[width=\linewidth]{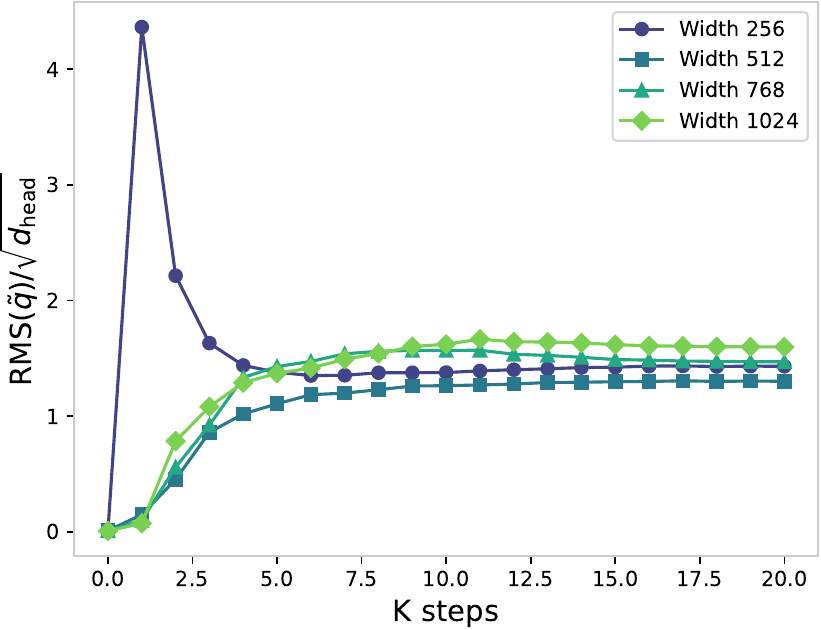}
        \caption{Our $\mu$P configuration}
    \end{subfigure}
    \caption{The curves for $\text{RMS}(\tilde{\qb})/\sqrt{d_\mathrm{head}}$ for SP, original $\mu$P configuration and our $\mu$P configuration.}
    \label{fig:q_hat_x_sqrtd}
\end{figure*}

\begin{figure*}[h]
    \centering
    \begin{subfigure}[b]{0.32\linewidth}
        \centering
        \includegraphics[width=\linewidth]{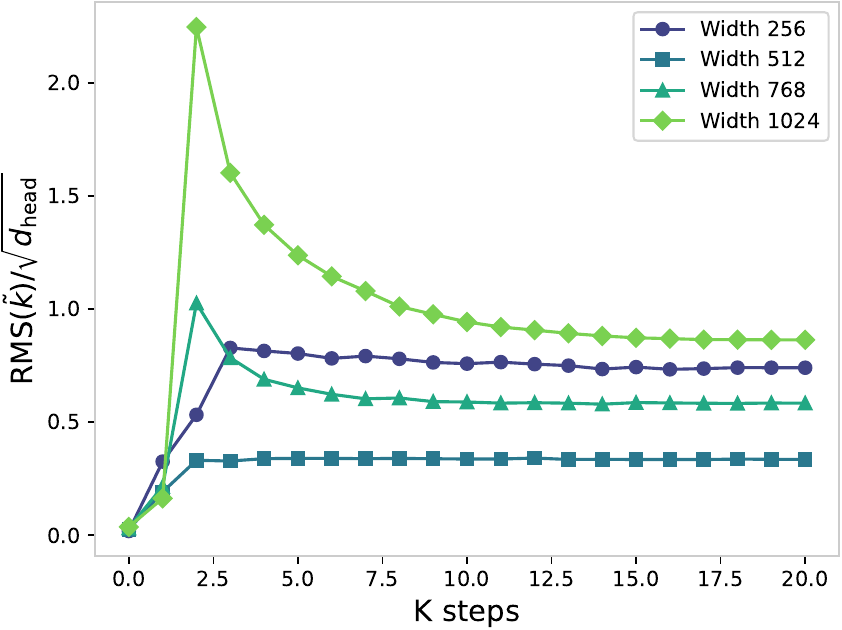}
        \caption{Standard Parametrization (SP)}
    \end{subfigure}
    \hfill
    \begin{subfigure}[b]{0.32\linewidth}
        \centering
        \includegraphics[width=\linewidth]{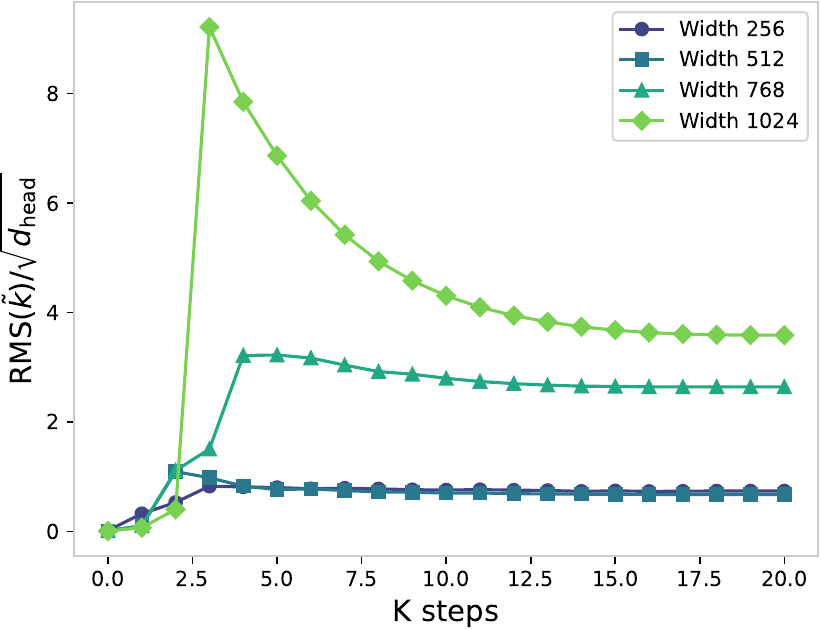}
        \caption{Original $\mu$P configuration}
    \end{subfigure}
    \hfill
    \begin{subfigure}[b]{0.32\linewidth}
        \centering
        \includegraphics[width=\linewidth]{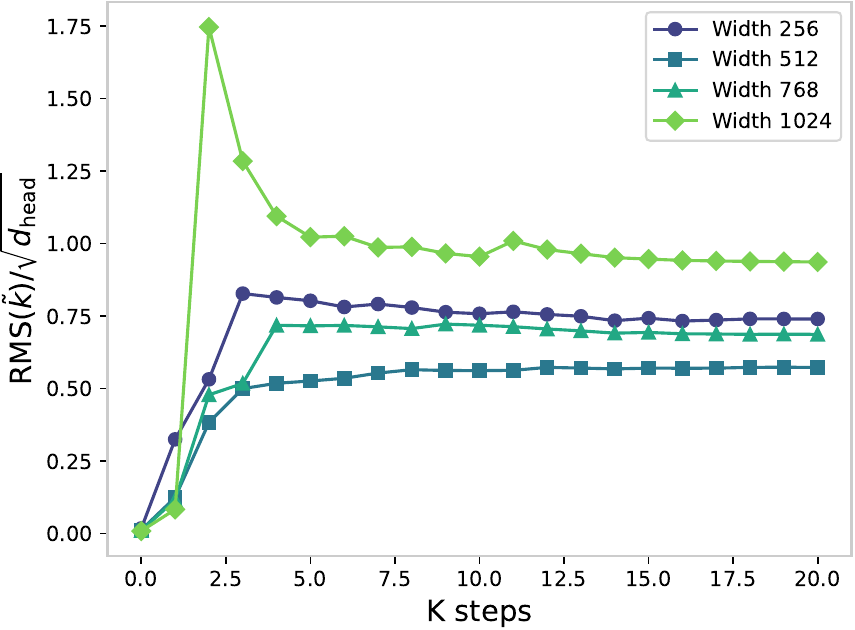}
        \caption{Our $\mu$P configuration}
    \end{subfigure}
    \caption{The curves for $\text{RMS}(\tilde{\kb})/\sqrt{d_\mathrm{head}}$ for SP, original $\mu$P configuration and our $\mu$P configuration.}
    \label{fig:k_hat_x_sqrtd}
\end{figure*}
\section{Dynamics analysis of $\mu$P for SGD}
In Sections~\ref{sec:mup_analysis} and~\ref{sec:mup_sgd}, we derived the $\mu$P formulation for the Gated Delta Network optimized by SGD by propagating coordinate-size estimates through the forward and backward passes. To corroborate these theoretical derivations, we embed diagnostic probes into our training framework. We track the internal activations, gradients, and parameter updates of models across different widths ($d\in\{256,512,768,1024\}$) during the pre-training phase with the same optimal learning rate of 0.4.

\subsection{Verification of Forward Pass Coordinate Sizes}
In Section~\ref{sec:mup_analysis}, we established that the query and key vectors before L2-normalization layer, $\tilde{\qb}_t$ and $\tilde{\kb}_t$, possess $\Theta(1/\sqrt{d})$ coordinate sizes. Figures~\ref{fig:q_hat_x_sqrtd} and~\ref{fig:k_hat_x_sqrtd} plot the empirical quantities $\tilde{\qb}/\sqrt{d_\mathrm{head}}\times\sqrt{d}$ and $\tilde{\kb}/\sqrt{d_\mathrm{head}}\times\sqrt{d}$ measured across the sampled layers, where we omitted $d=768$ due to instability across all the configurations.

As demonstrated, the scaled quantities is approximately constant across varying widths under our $\mu$P configuration, which does not hold for SP and original $\mu$P. This directly substantiates our derivations and validates that the recurrent state $\Sbb_t$ operates under stable variance conditions.

\begin{figure*}[h]
    \centering
    \begin{subfigure}[b]{0.32\linewidth}
        \centering
        \includegraphics[width=\linewidth]{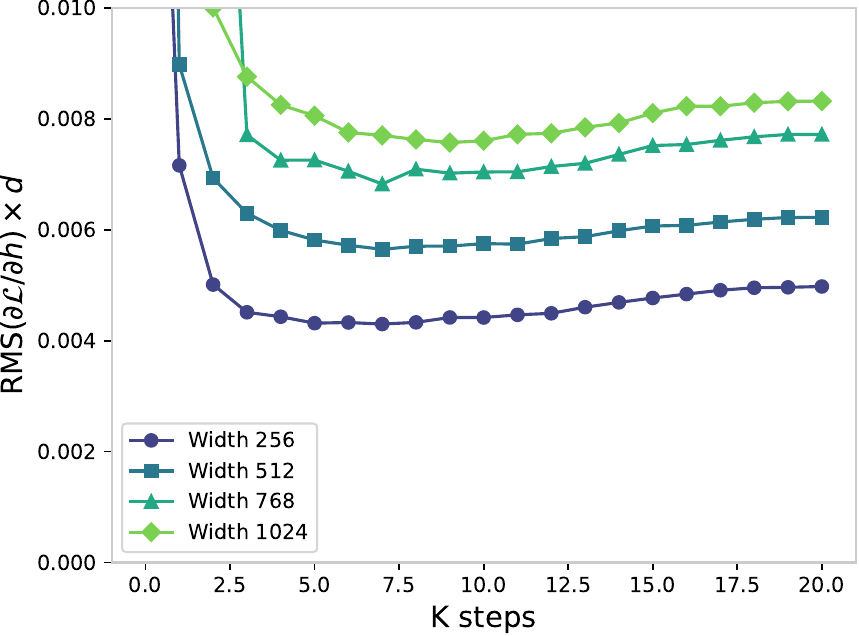}
        \caption{Standard Parametrization (SP)}
    \end{subfigure}
    \hfill
    \begin{subfigure}[b]{0.32\linewidth}
        \centering
        \includegraphics[width=\linewidth]{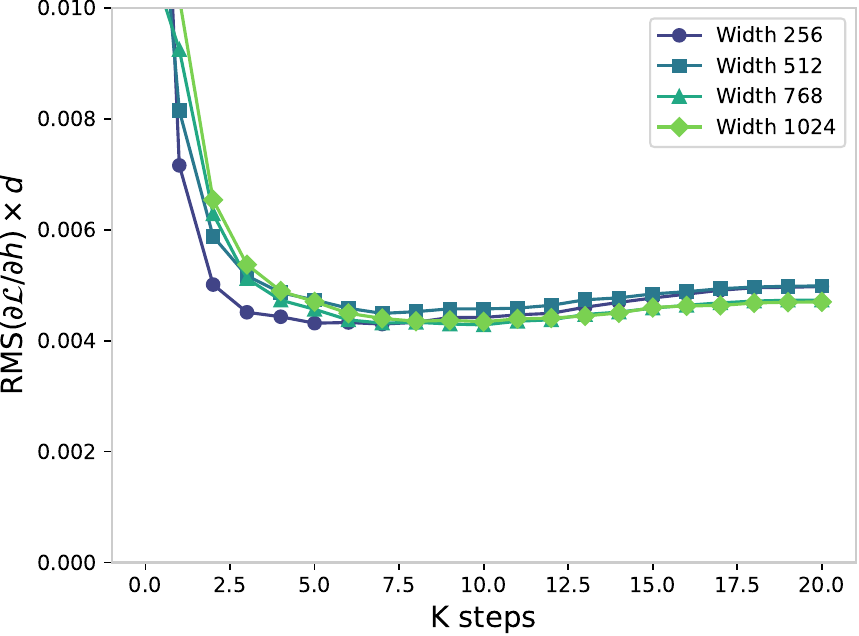}
        \caption{Original $\mu$P configuration}
    \end{subfigure}
    \hfill
    \begin{subfigure}[b]{0.32\linewidth}
        \centering
        \includegraphics[width=\linewidth]{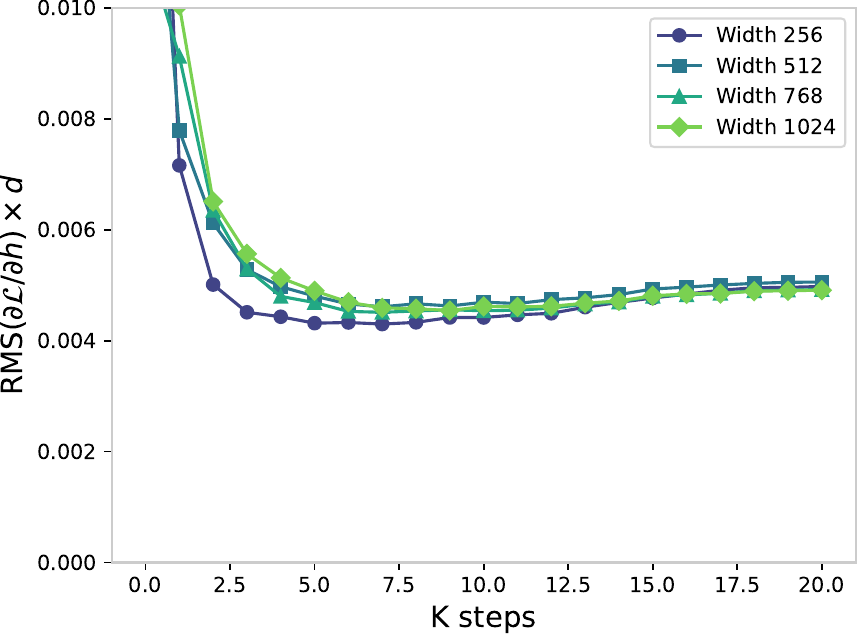}
        \caption{Our $\mu$P configuration}
    \end{subfigure}
    \caption{The curves for $\text{RMS}(\partial\mathcal{L}/\partial \hb)\times d$ for SP, original $\mu$P configuration and our $\mu$P configuration, where $h$ is the hidden state before each layer.}
    \label{fig:grad_hidden_rms}
\end{figure*}

\subsection{Verification of Backward Pass Gradient Scaling}
We also inspect the gradient scaling in the dynamics in backward pass. In Figure~\ref{fig:grad_hidden_rms}, we deduced that under our formulation and original $\mu$P configuration, the gradient for the hidden states $\partial \mathcal{L} / \partial \hb$ strictly follows a $\Theta(1/d)$ coordinate size, whereas standard parametrization forces it to $\Theta(1/\sqrt{d})$, leading to scaling instability.

\begin{figure*}[h]
    \centering
    \begin{subfigure}[b]{0.32\linewidth}
        \centering
        \includegraphics[width=\linewidth]{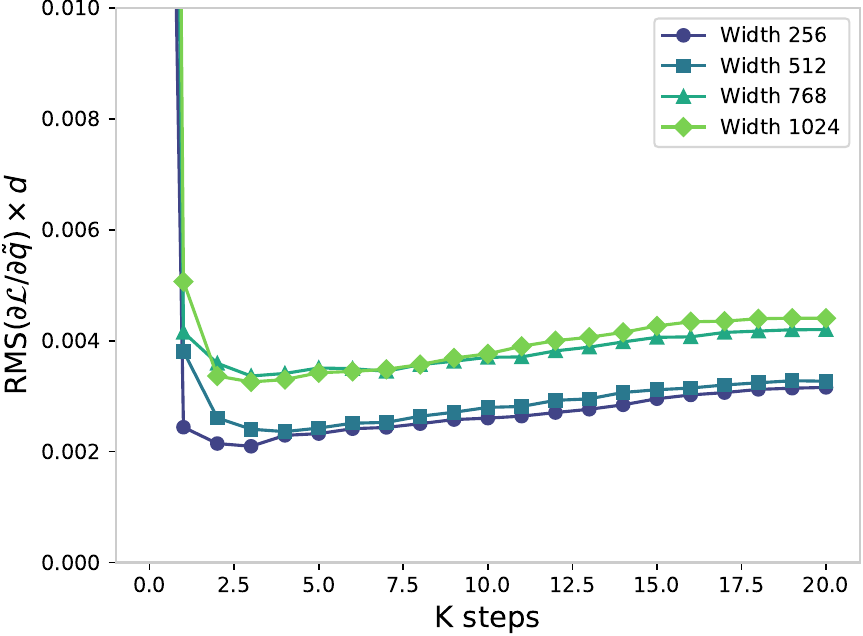}
        \caption{Standard Parametrization (SP)}
    \end{subfigure}
    \hfill
    \begin{subfigure}[b]{0.32\linewidth}
        \centering
        \includegraphics[width=\linewidth]{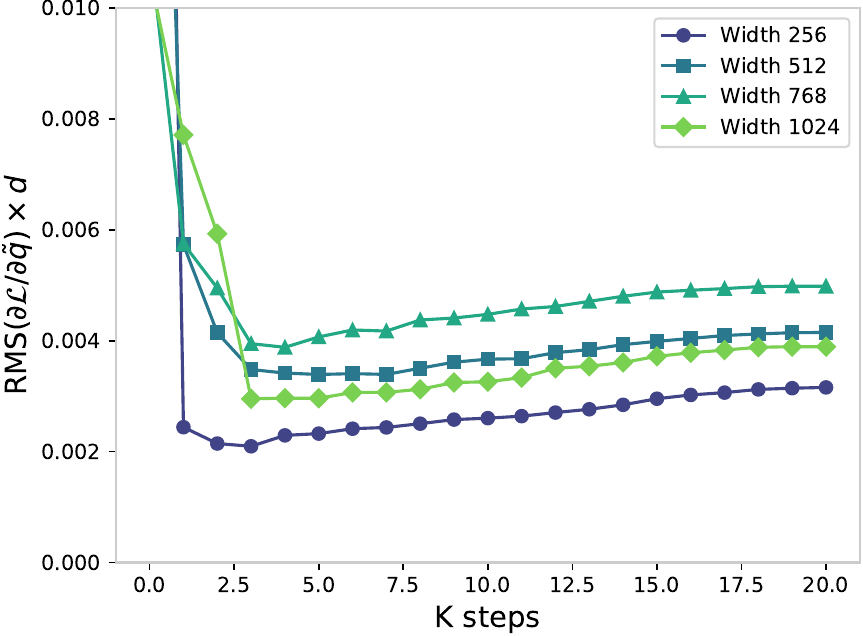}
        \caption{Original $\mu$P configuration}
    \end{subfigure}
    \hfill
    \begin{subfigure}[b]{0.32\linewidth}
        \centering
        \includegraphics[width=\linewidth]{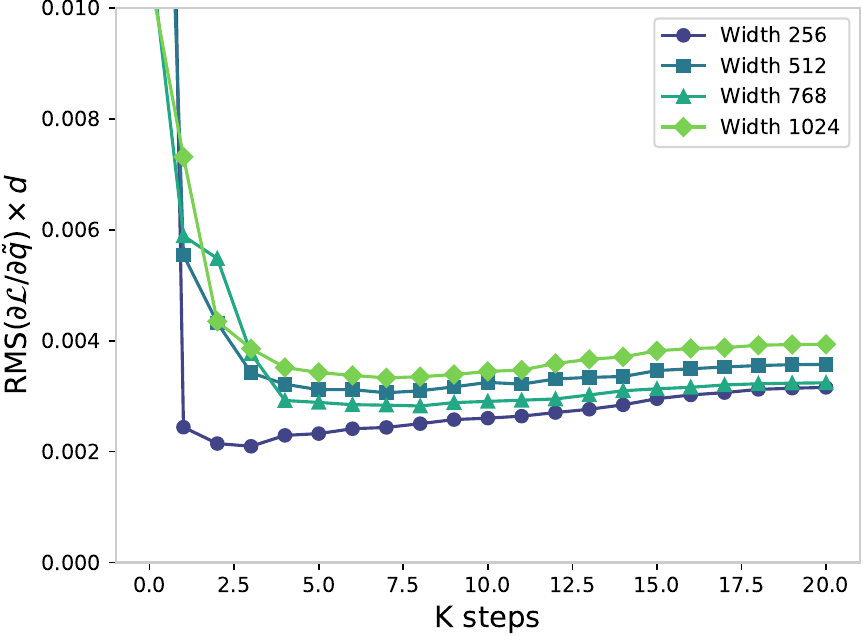}
        \caption{Our $\mu$P configuration}
    \end{subfigure}
    \caption{The curves for $\text{RMS}(\partial\mathcal{L}/\partial \tilde{\qb})\times d$ for SP, original $\mu$P configuration and our $\mu$P configuration.}
    \label{fig:grad_q_rms}
\end{figure*}

\begin{figure*}[h]
    \centering
    \begin{subfigure}[b]{0.32\linewidth}
        \centering
        \includegraphics[width=\linewidth]{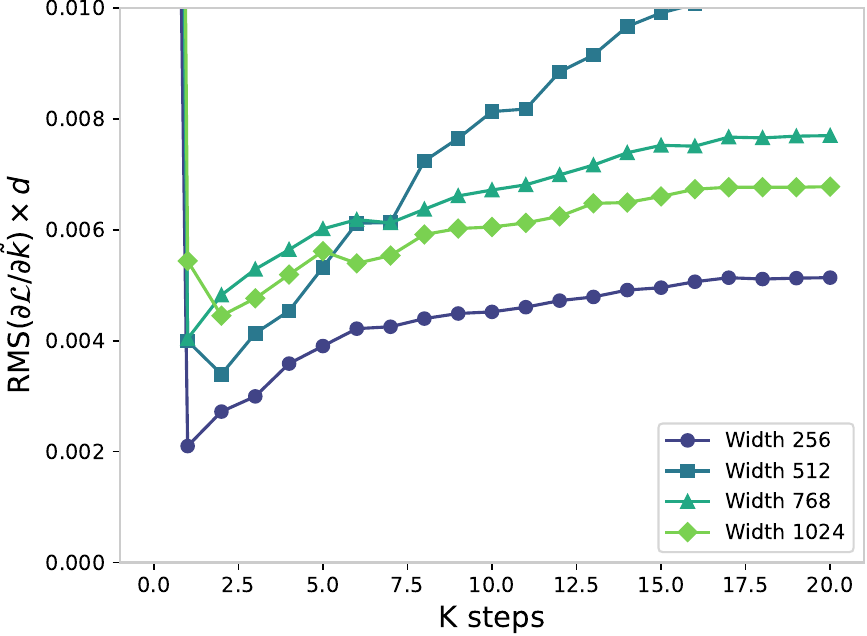}
        \caption{Standard Parametrization (SP)}
    \end{subfigure}
    \hfill
    \begin{subfigure}[b]{0.32\linewidth}
        \centering
        \includegraphics[width=\linewidth]{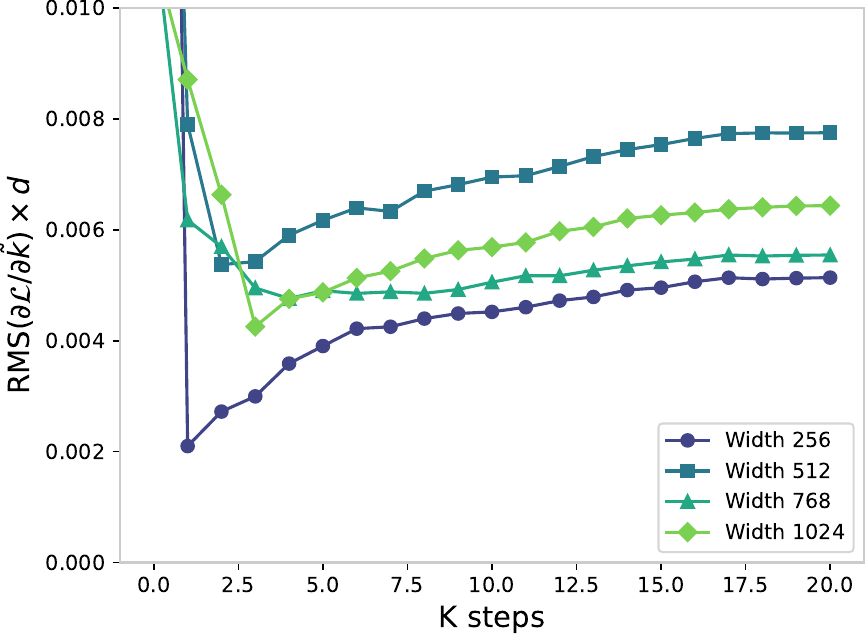}
        \caption{Original $\mu$P configuration}
    \end{subfigure}
    \hfill
    \begin{subfigure}[b]{0.32\linewidth}
        \centering
        \includegraphics[width=\linewidth]{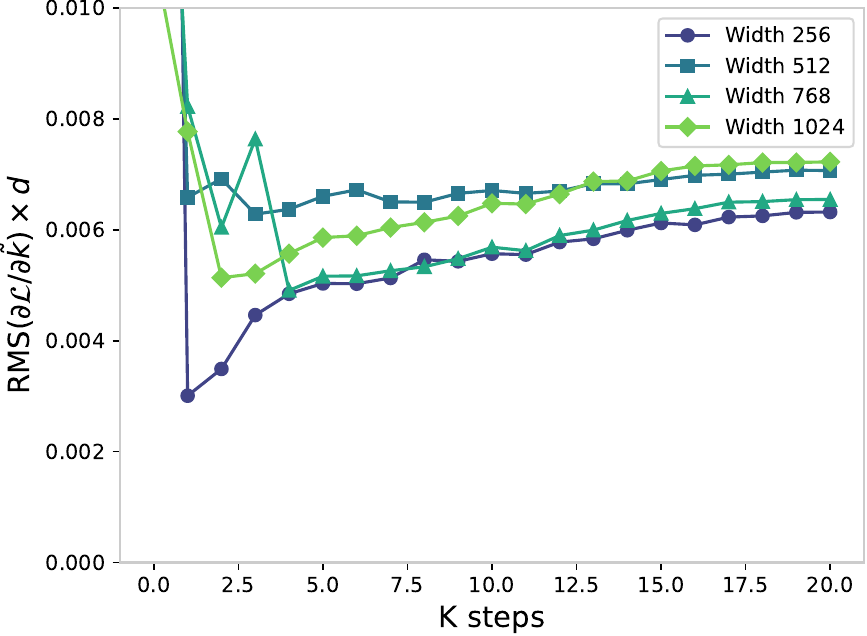}
        \caption{Our $\mu$P configuration}
    \end{subfigure}
    \caption{The curves for $\text{RMS}(\partial\mathcal{L}/\partial \tilde{\kb})\times d$ for SP, original $\mu$P configuration and our $\mu$P configuration.}
    \label{fig:grad_k_rms}
\end{figure*}

Moreover, in Figures~\ref{fig:grad_q_rms} and~\ref{fig:grad_k_rms}, we observed that both $\partial \mathcal{L} / \partial \tilde{\qb}$ and $\partial \mathcal{L} / \partial \tilde{\kb}$ also follows a $\Theta(1/d)$ coordinate size in our configuration, while original $\mu$P deviates from it, making the training less stable and transfer not perfect.


\begin{figure*}
    \centering
    \includegraphics[width=0.5\linewidth]{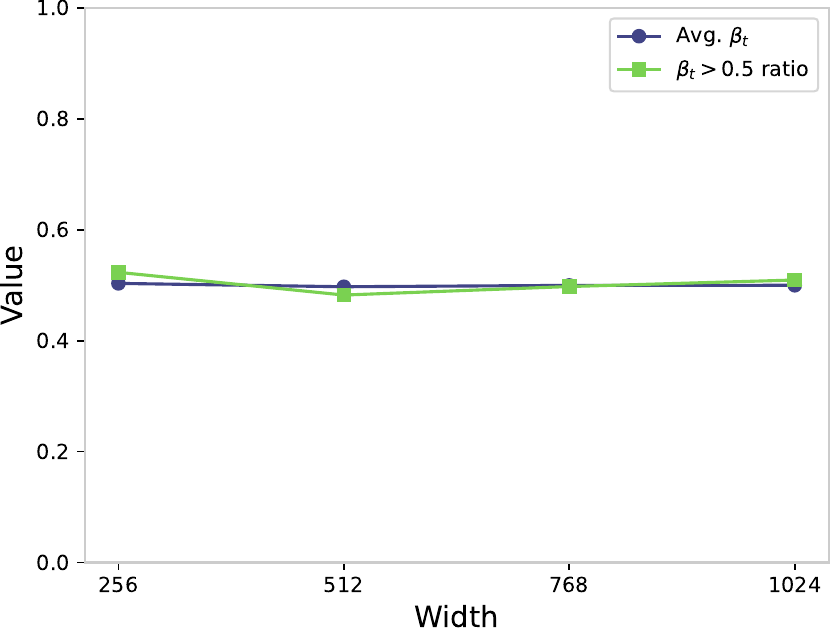}
    \caption{The average of $\beta_t$ and the ratio for strong-writing $\beta_t$ ($\beta_t>0.5$) for GDN models trained with SGD in our $\mu$P configuration with different widths.}
    \label{fig:beta_stat}
\end{figure*}
\begin{figure*}[htb!]
    \centering
    \begin{subfigure}[b]{0.32\linewidth}
        \centering
        \includegraphics[width=\linewidth]{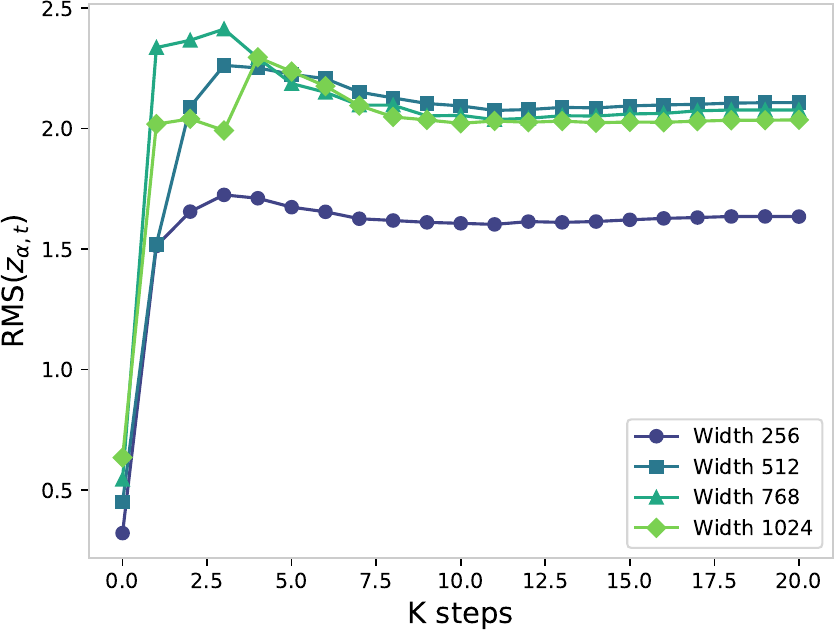}
        \caption{Standard Parametrization (SP)}
    \end{subfigure}
    \hfill
    \begin{subfigure}[b]{0.32\linewidth}
        \centering
        \includegraphics[width=\linewidth]{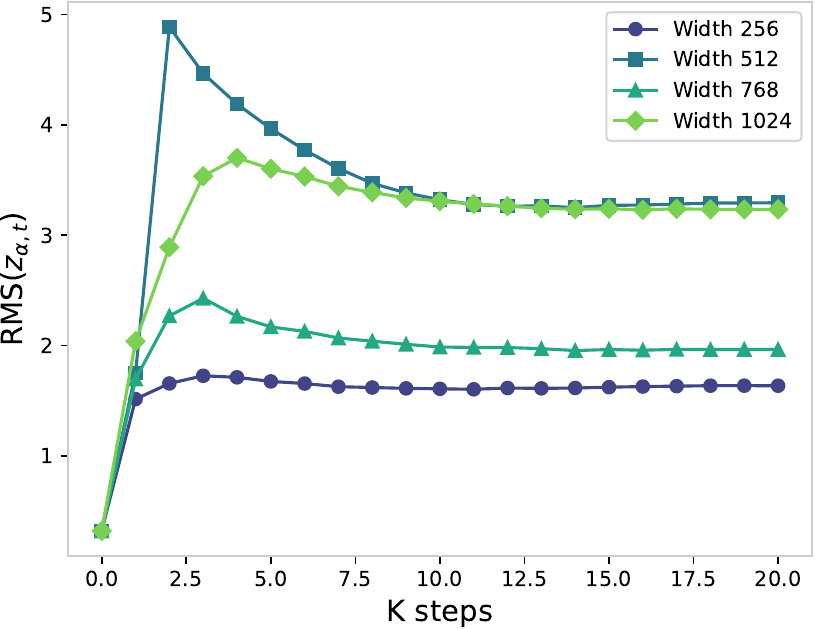}
        \caption{Original $\mu$P configuration}
    \end{subfigure}
    \hfill
    \begin{subfigure}[b]{0.32\linewidth}
        \centering
        \includegraphics[width=\linewidth]{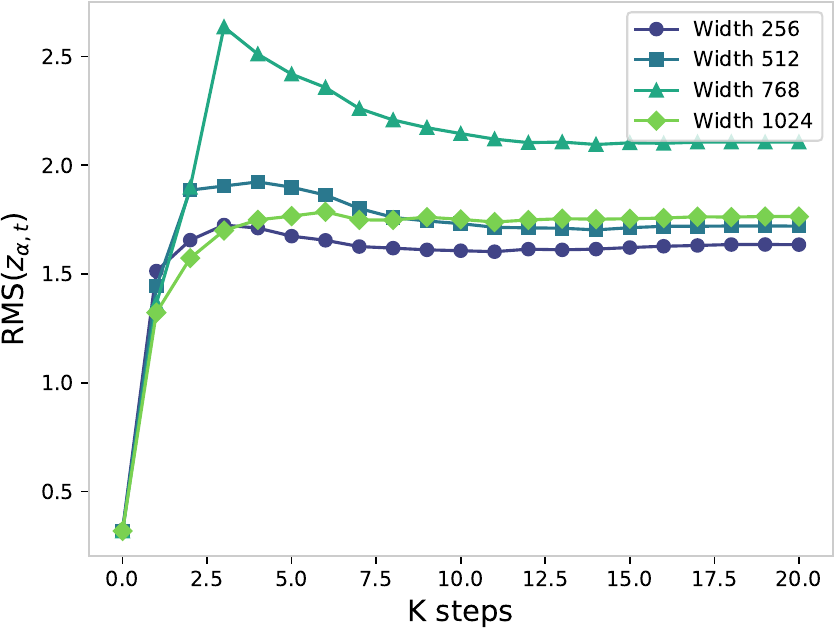}
        \caption{Our $\mu$P configuration}
    \end{subfigure}
    \caption{The curves for $\text{RMS}(z_{\alpha,t})$ for SP, original $\mu$P configuration and our $\mu$P configuration.}
    \label{fig:a_proj_output_std}
\end{figure*}
\begin{figure*}[h]
    \centering
    \begin{subfigure}[b]{0.32\linewidth}
        \centering
        \includegraphics[width=\linewidth]{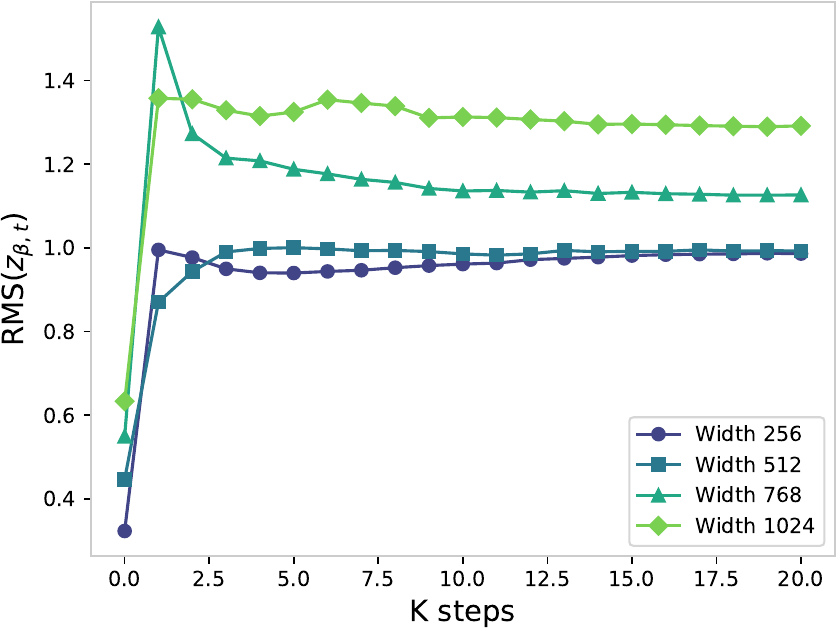}
        \caption{Standard Parametrization (SP)}
    \end{subfigure}
    \hfill
    \begin{subfigure}[b]{0.32\linewidth}
        \centering
        \includegraphics[width=\linewidth]{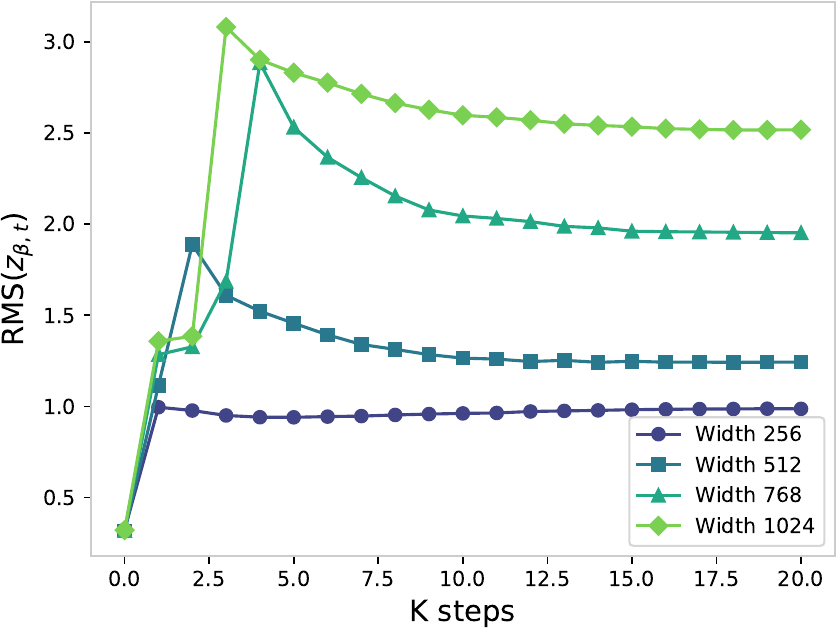}
        \caption{Original $\mu$P configuration}
    \end{subfigure}
    \hfill
    \begin{subfigure}[b]{0.32\linewidth}
        \centering
        \includegraphics[width=\linewidth]{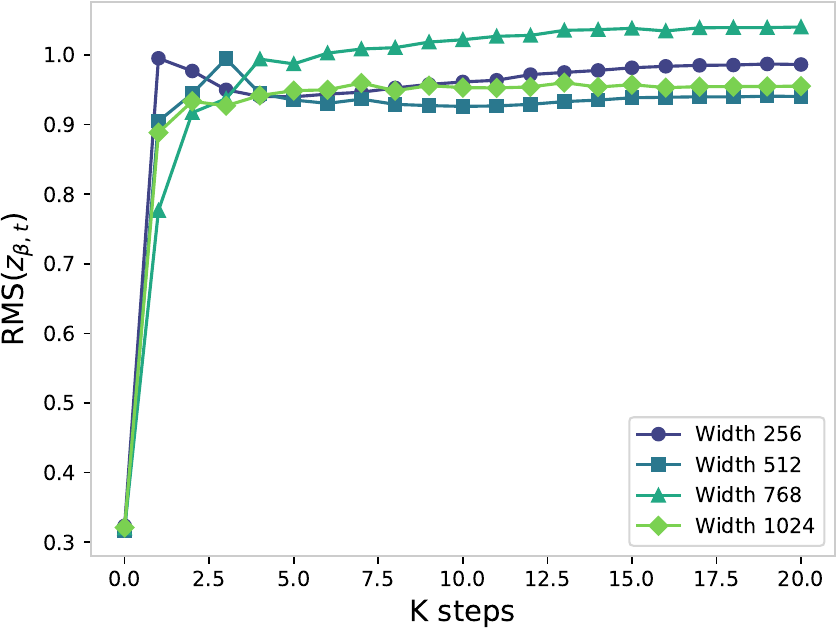}
        \caption{Our $\mu$P configuration}
    \end{subfigure}
    \caption{The curves for average standard deviation of $\text{RMS}(z_{\beta,t})$ for SP, original $\mu$P configuration and our $\mu$P configuration.}
    \label{fig:b_proj_output_std}
\end{figure*}
\subsection{Stability of the Gating Dynamics}
Our theoretical approximations in Section~\ref{sec:mup_analysis} assume that the data-dependent gating scalars ($\alpha_t$ and $\beta_t$) do not saturate into trivial states (e.g., vanishing completely or remaining strictly 1.0). 

By logging the runtime statistics of the write strength $\beta = \sigma(W_\beta x_t)$ in Figure~\ref{fig:beta_stat}, we observe that its expected value remains centered around 0.5 across all model widths, with a significant proportion of tokens actively triggering strong writes. Additionally, in Figures~\ref{fig:a_proj_output_std} and~\ref{fig:b_proj_output_std} ($d=768$ is omitted for $z_{\beta,t}$ for instability across all the configurations), the standard deviation of the pre-activations $z_{\alpha,t}=\Wb_\alpha \xb_t+b$ and $z_{\beta,t}=\Wb_\beta\xb_t$ remains $\Theta(1)$. These observations validate our first-order approximations and ensure that the recurrent memory updates effectively capture long-range dependencies without collapsing as the model scales.

\begin{figure*}[h]
    \centering
    \begin{subfigure}[b]{0.32\linewidth}
        \centering
        \includegraphics[width=\linewidth]{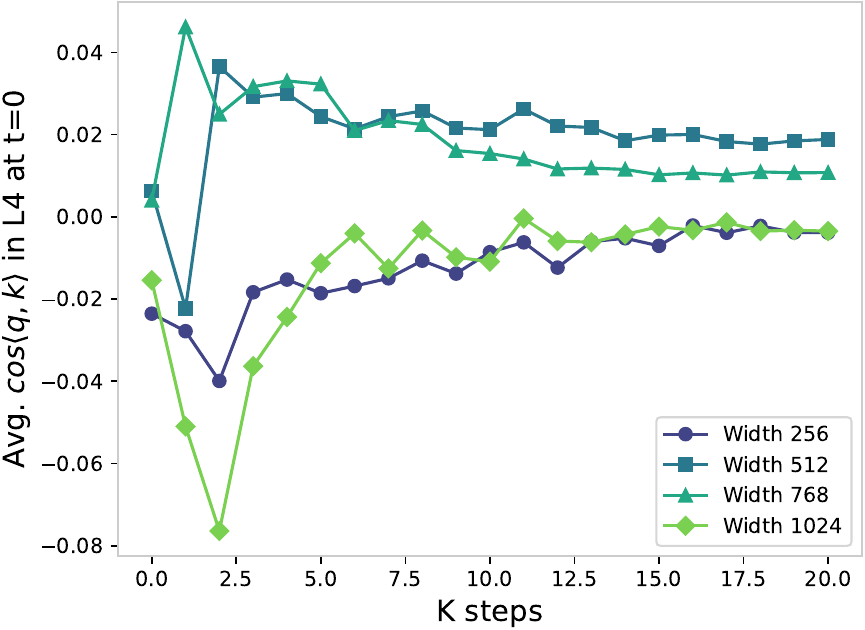}
        \caption{$\cos\langle \qb_t,\kb_t\rangle$ at $t=0$}
    \end{subfigure}
    \begin{subfigure}[b]{0.32\linewidth}
        \centering
        \includegraphics[width=\linewidth]{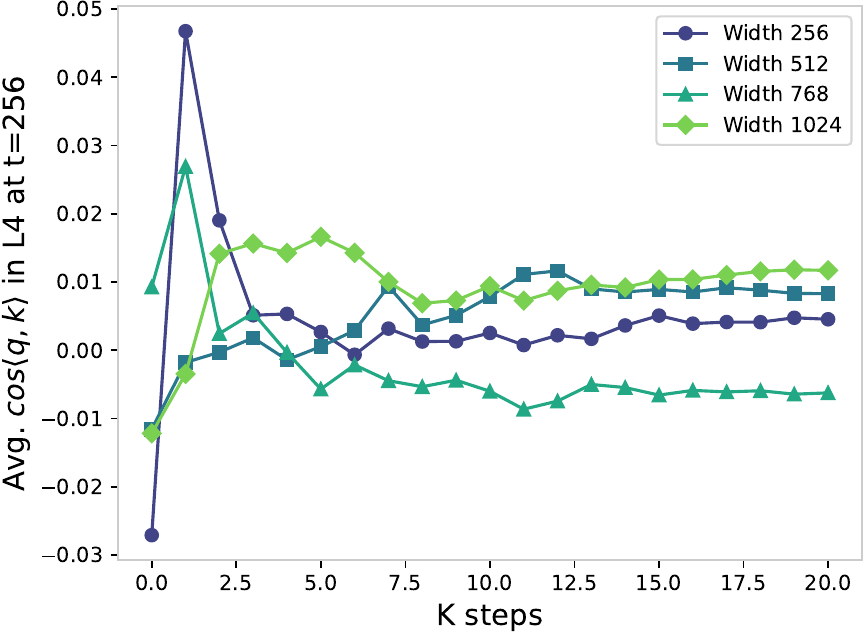}
        \caption{$\cos\langle \qb_t,\kb_t\rangle$ at $t=256$}
    \end{subfigure}
    \begin{subfigure}[b]{0.32\linewidth}
        \centering
        \includegraphics[width=\linewidth]{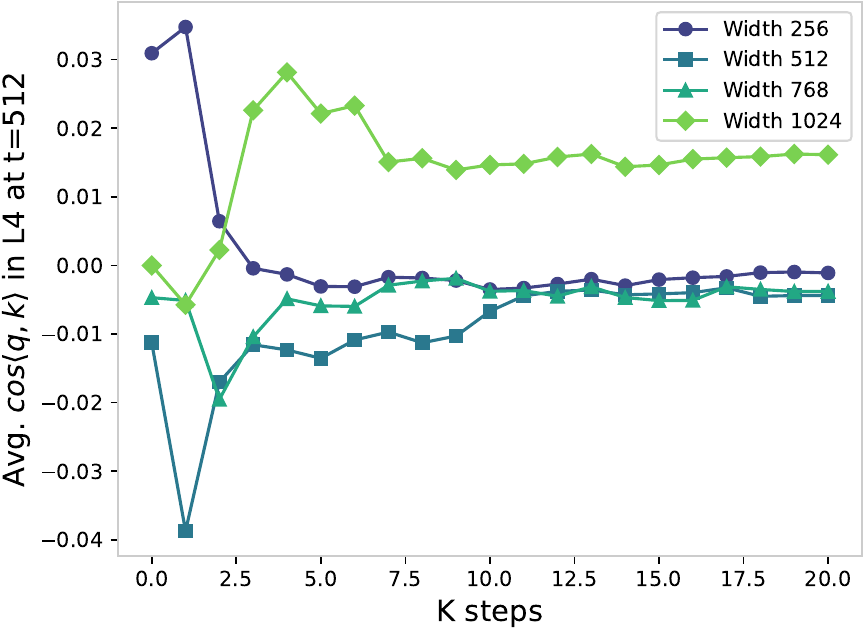}
        \caption{$\cos\langle \qb_t,\kb_t\rangle$ at $t=512$}
    \end{subfigure}
    \begin{subfigure}[b]{0.32\linewidth}
        \centering
        \includegraphics[width=\linewidth]{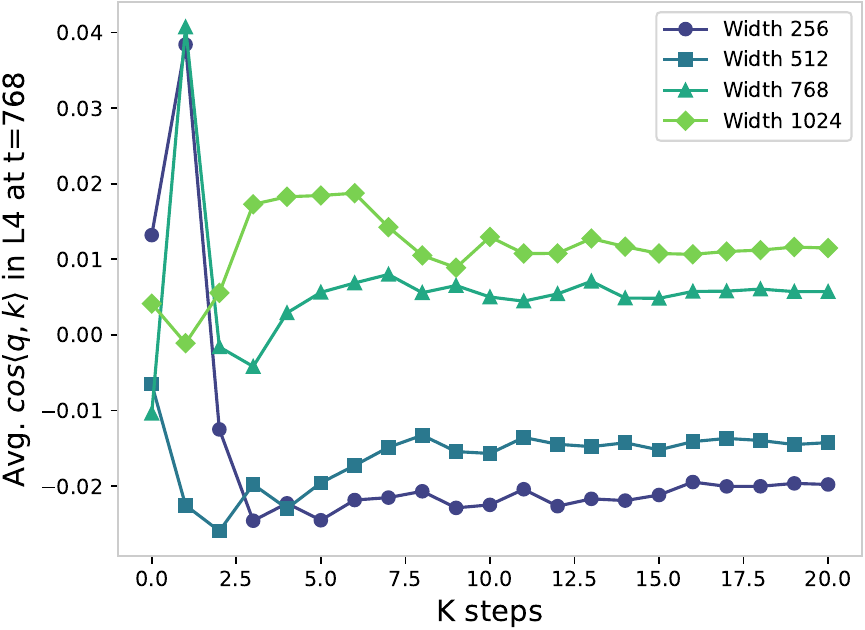}
        \caption{$\cos\langle \qb_t,\kb_t\rangle$ at $t=768$}
    \end{subfigure}
    \begin{subfigure}[b]{0.32\linewidth}
        \centering
        \includegraphics[width=\linewidth]{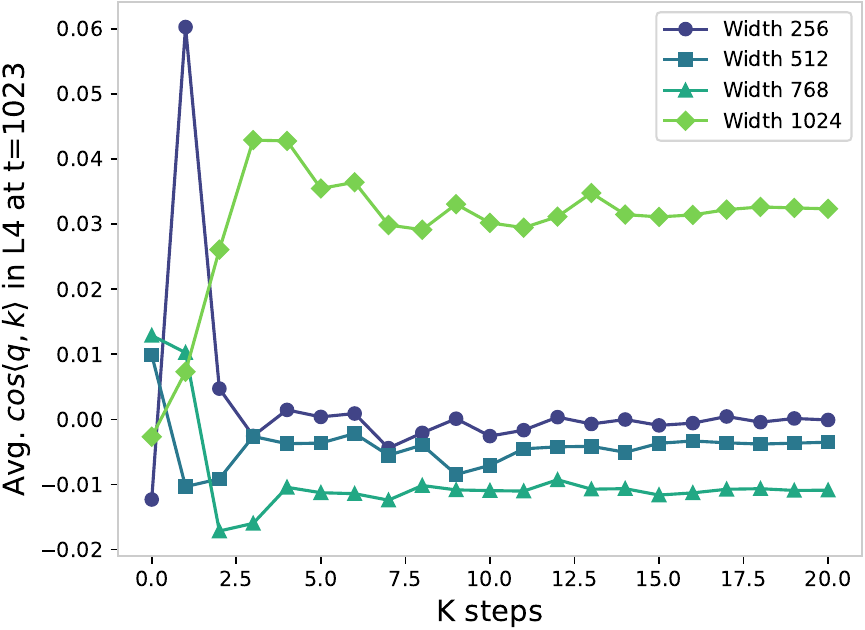}
        \caption{$\cos\langle \qb_t,\kb_t\rangle$ at $t=1023$}
    \end{subfigure}
    \caption{The curves for $\cos\langle \qb_t,\kb_t\rangle$ in our $\mu$P configuration for SGD at different $t$ values at 4-th layer.}
    \label{fig:qk_cos}
\end{figure*}
\subsection{Detecting the dynamics of state spaces}
Finally, we depict the dynamics of state spaces of GDN when $t$ increases. We plotted the $\cos\langle \qb_t,\kb_t\rangle$ given a certain input in our $\mu$P configuration at different $t$ values at 4-th layer in Figure~\ref{fig:qk_cos}. It can be seen that all the values are around 0, showing that independent assumption of $\qb_t$ and $\kb_t$ still holds for GDN. And during the training process, the value also converges, showing that in our configuration, the model can indeed converge to a stable state.
\section{Dynamic analysis of AdamW}
In previous section, we show the plots of various dynamic probes of the model trained with SGD. Here we provide additional analysis of the dynamics of Gated Delta Net trained with AdamW optimizer across different widths among $d\in\{256,512,1024,1536\}$ with the same optimal learning rate of $8\times 10^{-3}$.
\begin{figure*}[h]
    \centering
    \begin{subfigure}[b]{0.45\linewidth}
        \centering
        \includegraphics[width=\linewidth]{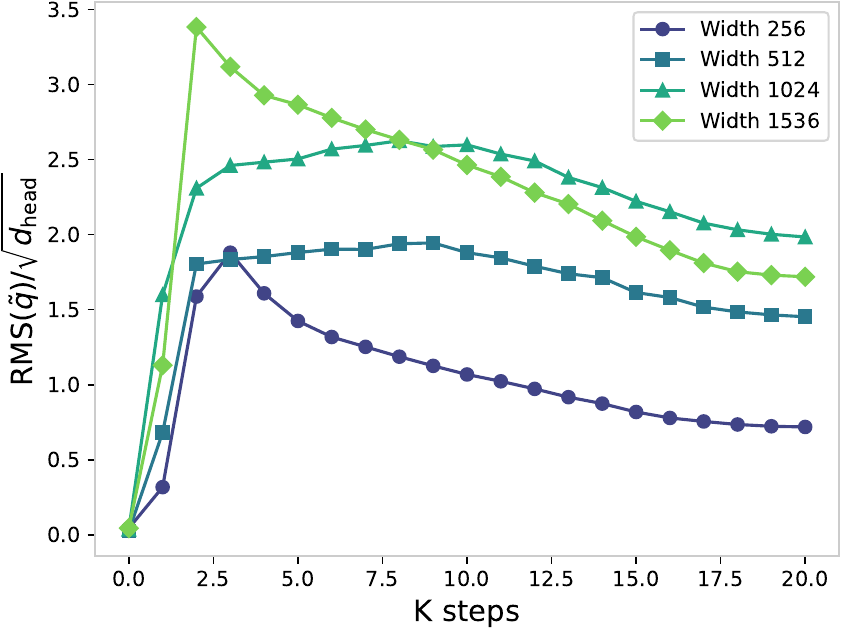}
        \caption{Standard Parametrization (SP)}
    \end{subfigure}
    \begin{subfigure}[b]{0.45\linewidth}
        \centering
        \includegraphics[width=\linewidth]{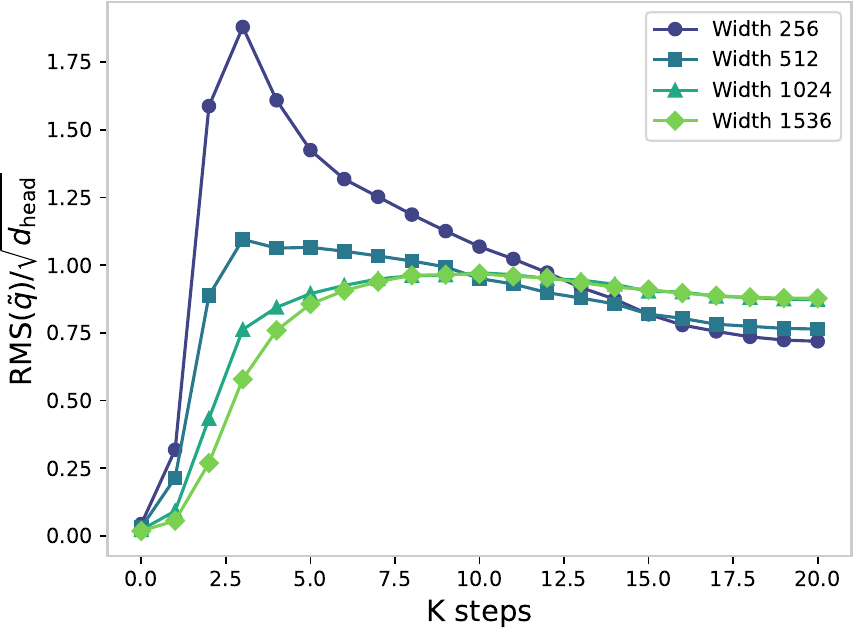}
        \caption{Original $\mu$P configuration}
    \end{subfigure}
    \caption{The curves for $\text{RMS}(\tilde{\qb})/\sqrt{d_\mathrm{head}}$ for SP and $\mu$P configuration with AdamW optimizer.}
    \label{fig:q_hat_x_sqrtd_adamw}
\end{figure*}

\begin{figure*}[h]
    \centering
    \begin{subfigure}[b]{0.45\linewidth}
        \centering
        \includegraphics[width=\linewidth]{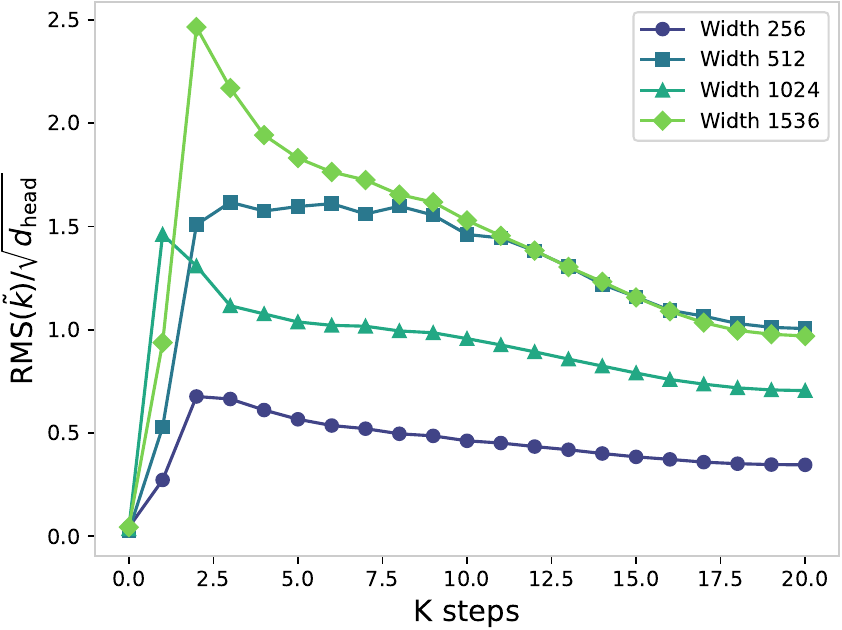}
        \caption{Standard Parametrization (SP)}
    \end{subfigure}
    \begin{subfigure}[b]{0.45\linewidth}
        \centering
        \includegraphics[width=\linewidth]{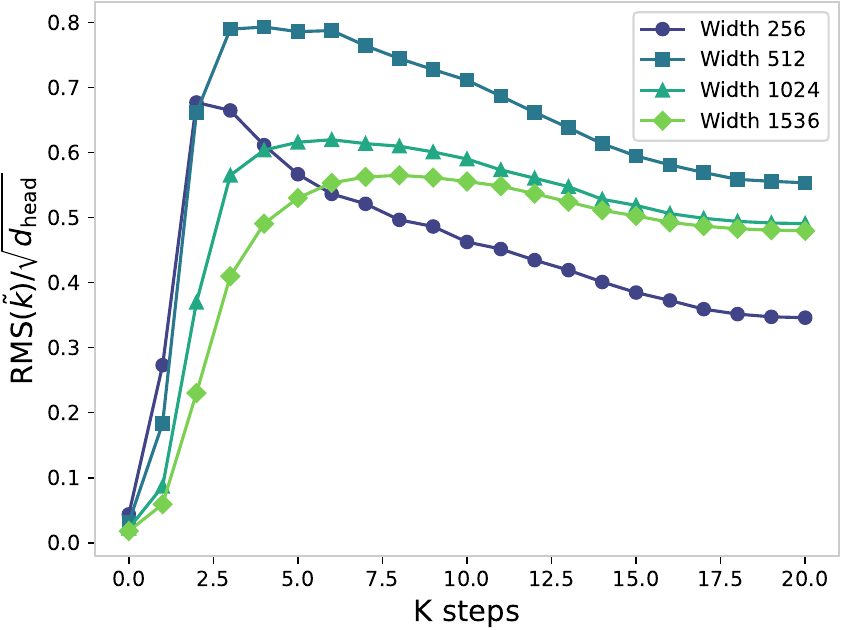}
        \caption{$\mu$P configuration}
    \end{subfigure}
    \caption{The curves for $\text{RMS}(\tilde{\kb})/\sqrt{d_\mathrm{head}}$ for SP and $\mu$P configuration with AdamW optimizer.}
    \label{fig:k_hat_x_sqrtd_adamw}
\end{figure*}
\subsection{Verification of Forward Pass Coordinate Sizes}
In Section~\ref{sec:mup_analysis}, we demonstrated that $\tilde{\qb}_t$ and $\tilde{\kb}_t$ have $\Theta(1/\sqrt{d})$ coordinate sizes. We display the empirical quantities $\tilde{\qb}/\sqrt{d_\mathrm{head}}\times\sqrt{d}$ and $\tilde{\kb}/\sqrt{d_\mathrm{head}}\times\sqrt{d}$ measured across the sampled layers in Figures~\ref{fig:q_hat_x_sqrtd_adamw} and~\ref{fig:k_hat_x_sqrtd_adamw}.

We can observe that the scaled quantities is approximately constant across varying widths under our $\mu$P configuration, which does not hold for SP. This directly substantiates our derivations and validates that the recurrent state $\Sbb_t$ operates under stable variance conditions.

\begin{figure*}[h]
    \centering
    \begin{subfigure}[b]{0.45\linewidth}
        \centering
        \includegraphics[width=\linewidth]{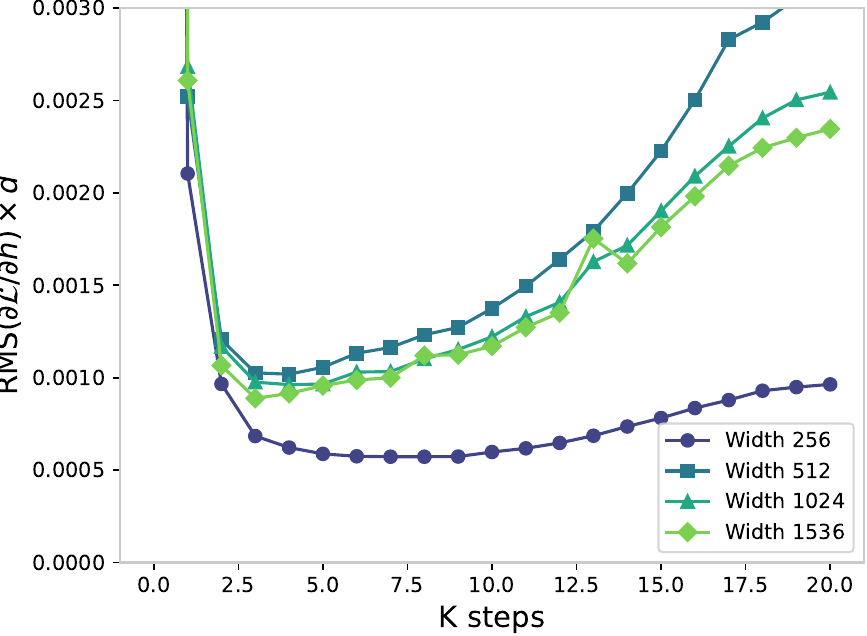}
        \caption{Standard Parametrization (SP)}
    \end{subfigure}
    \begin{subfigure}[b]{0.45\linewidth}
        \centering
        \includegraphics[width=\linewidth]{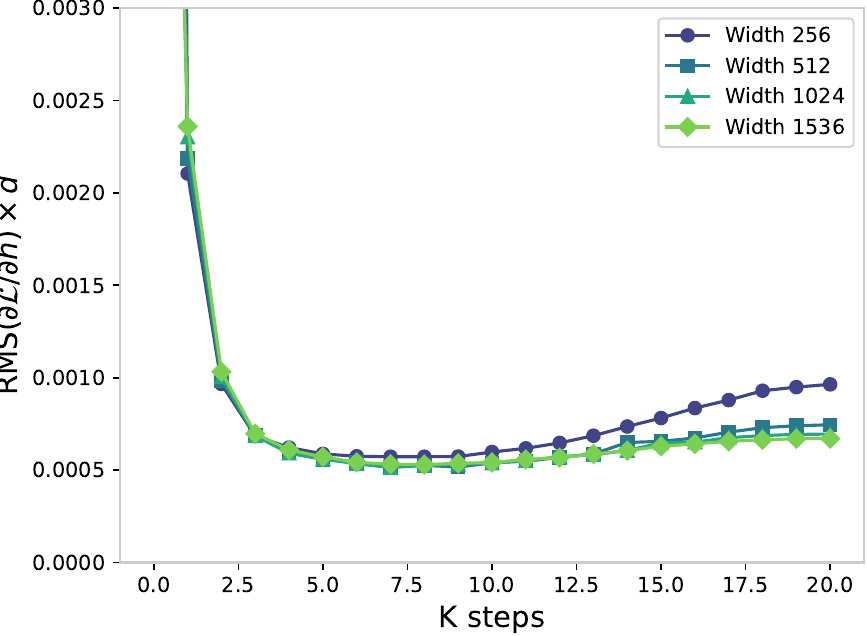}
        \caption{$\mu$P configuration}
    \end{subfigure}
    \caption{The curves for $\text{RMS}(\partial\mathcal{L}/\partial \hb)\times d$ for SP and $\mu$P configuration with AdamW optimizer, where $h$ is the hidden state before each layer.}
    \label{fig:grad_hidden_rms_adamw}
\end{figure*}

\subsection{Inspect of Backward Pass Gradient Scaling}
We also inspect the gradient scaling in the dynamics in backward pass. Although the normalization within Adam(W) with $\mu$P configuration ensures the $\Theta(1)$ update of hidden state, here we instead inspect the stability of the gradient across different layers. In Figure~\ref{fig:grad_hidden_rms_adamw}, we notice that under $\mu$P configuration, the gradient for the hidden states $\partial \mathcal{L} / \partial \hb$ is much more stabler than that with SP configuration.

\begin{figure*}
    \centering
    \includegraphics[width=0.5\linewidth]{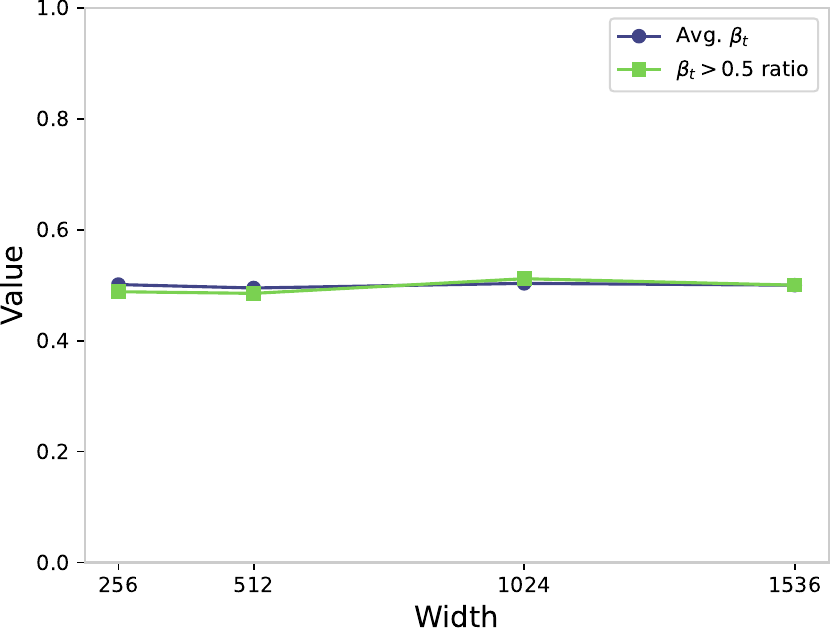}
    \caption{The average of $\beta_t$ and the ratio for strong-writing $\beta_t$ ($\beta_t>0.5$) for GDN models trained with AdamW in $\mu$P configuration with different widths.}
    \label{fig:beta_stat_adamw}
\end{figure*}
\begin{figure*}[htb!]
    \centering
    \begin{subfigure}[b]{0.45\linewidth}
        \centering
        \includegraphics[width=\linewidth]{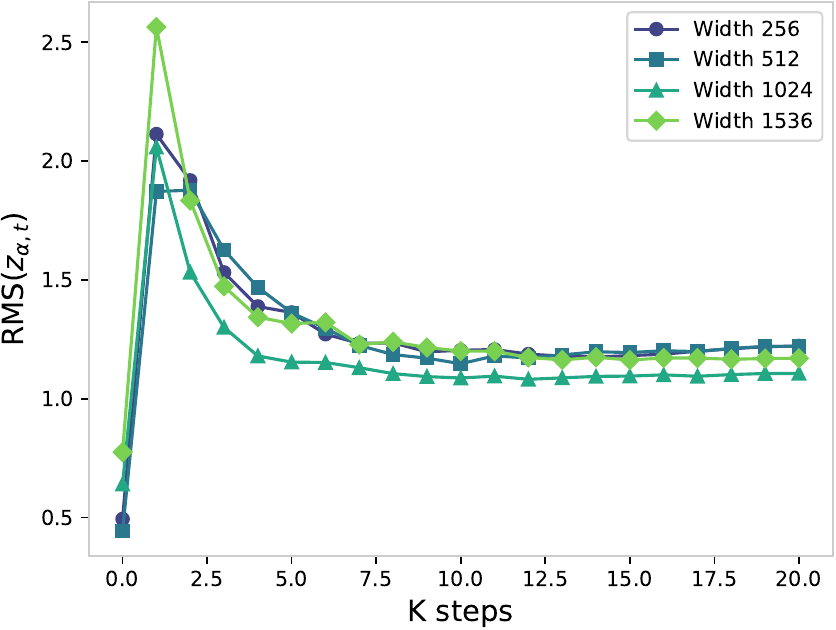}
        \caption{Standard Parametrization (SP)}
    \end{subfigure}
    \begin{subfigure}[b]{0.45\linewidth}
        \centering
        \includegraphics[width=\linewidth]{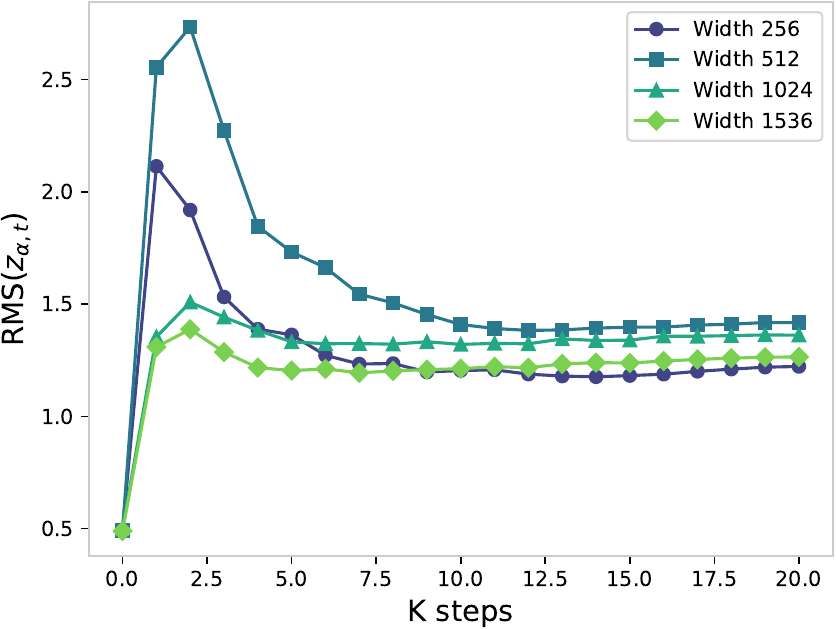}
        \caption{$\mu$P configuration}
    \end{subfigure}
    \caption{The curves for $\text{RMS}(z_{\alpha,t})$ for SP and $\mu$P configuration with AdamW optimizer.}
    \label{fig:a_proj_output_std_adamw}
\end{figure*}
\begin{figure*}[h]
    \centering
    \begin{subfigure}[b]{0.45\linewidth}
        \centering
        \includegraphics[width=\linewidth]{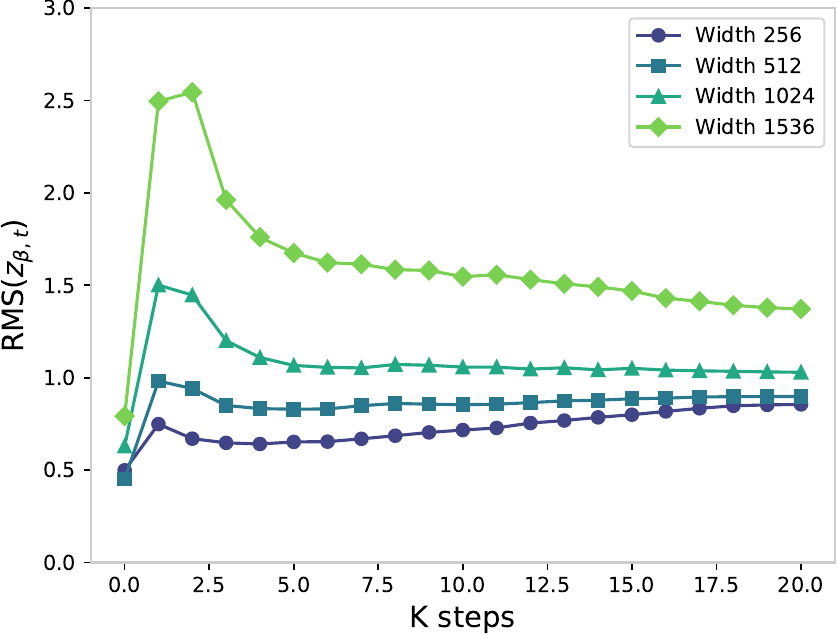}
        \caption{Standard Parametrization (SP)}
    \end{subfigure}
    \begin{subfigure}[b]{0.45\linewidth}
        \centering
        \includegraphics[width=\linewidth]{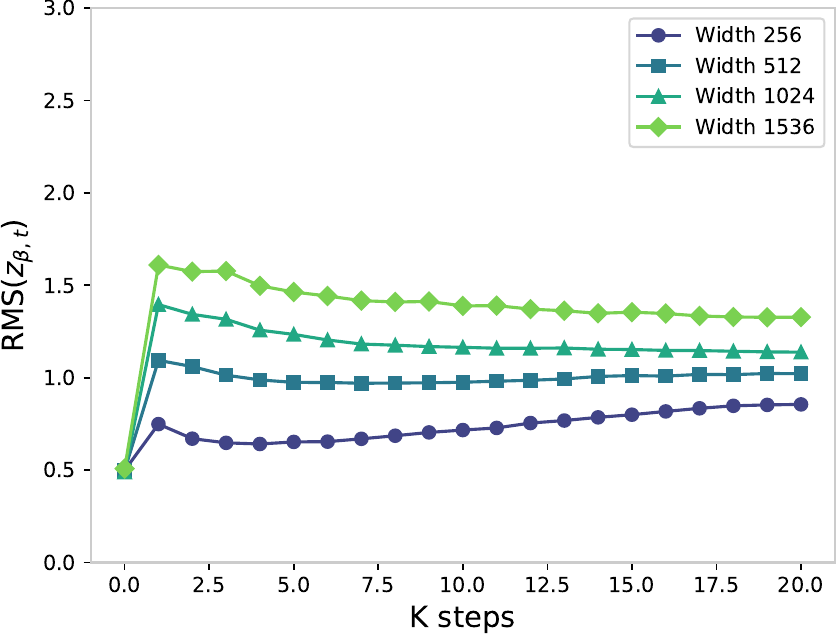}
        \caption{$\mu$P configuration}
    \end{subfigure}
    \caption{The curves for average standard deviation of $\text{RMS}(z_{\beta,t})$ for SP, original $\mu$P configuration and our $\mu$P configuration with AdamW optimizer.}
    \label{fig:b_proj_output_std_adamw}
\end{figure*}
\subsection{Stability of the Gating Dynamics}
In Section~\ref{sec:mup_analysis}, we assume that the data-dependent gating scalars ($\alpha_t$ and $\beta_t$) do not saturate into trivial states. In Figure~\ref{fig:beta_stat_adamw}, we observe that the expected value of $\beta_t$ remains around 0.5 across all model widths, with also a significant proportion of tokens actively triggering strong writes. Additionally, in Figures~\ref{fig:a_proj_output_std_adamw} and~\ref{fig:b_proj_output_std_adamw}, the standard deviation of the pre-activations $z_{\alpha,t}=\Wb_\alpha \xb_t+b$ and $z_{\beta,t}=\Wb_\beta\xb_t$ remains $\Theta(1)$. These observations validate our first-order approximations and ensure that the recurrent memory updates effectively capture long-range dependencies without collapsing as the model scales for AdamW scenarios.

\begin{figure*}[h]
    \centering
    \begin{subfigure}[b]{0.32\linewidth}
        \centering
        \includegraphics[width=\linewidth]{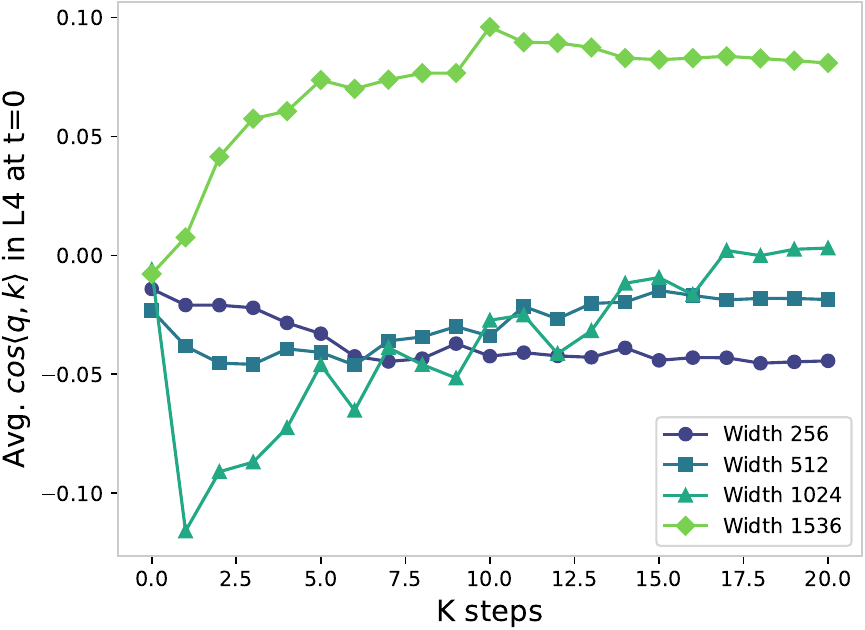}
        \caption{$\cos\langle \qb_t,\kb_t\rangle$ at $t=0$}
    \end{subfigure}
    \begin{subfigure}[b]{0.32\linewidth}
        \centering
        \includegraphics[width=\linewidth]{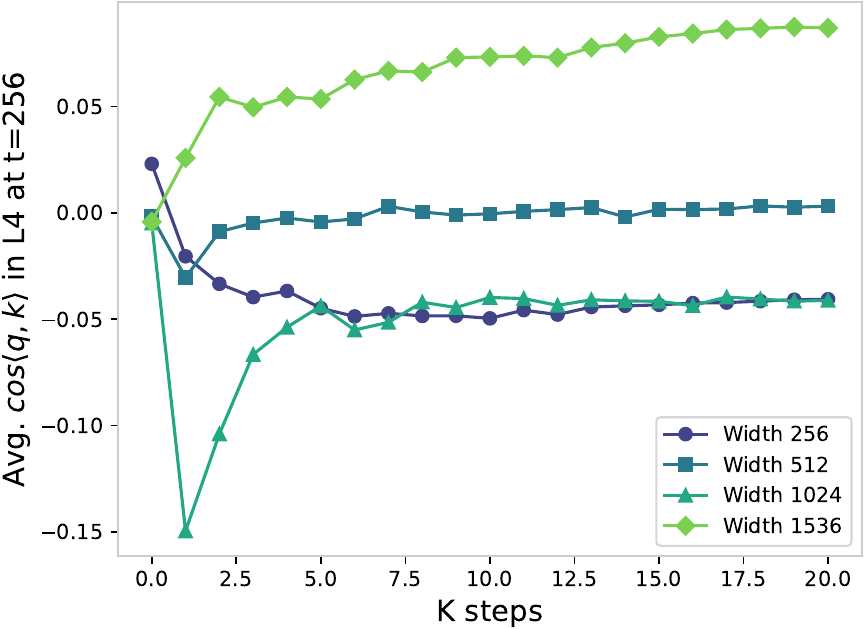}
        \caption{$\cos\langle \qb_t,\kb_t\rangle$ at $t=256$}
    \end{subfigure}
    \begin{subfigure}[b]{0.32\linewidth}
        \centering
        \includegraphics[width=\linewidth]{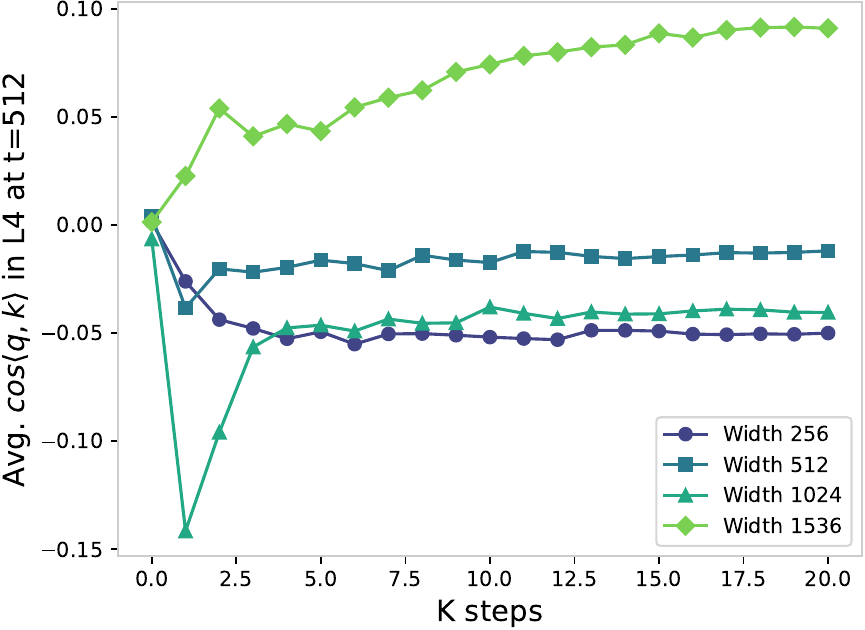}
        \caption{$\cos\langle \qb_t,\kb_t\rangle$ at $t=512$}
    \end{subfigure}
    \begin{subfigure}[b]{0.32\linewidth}
        \centering
        \includegraphics[width=\linewidth]{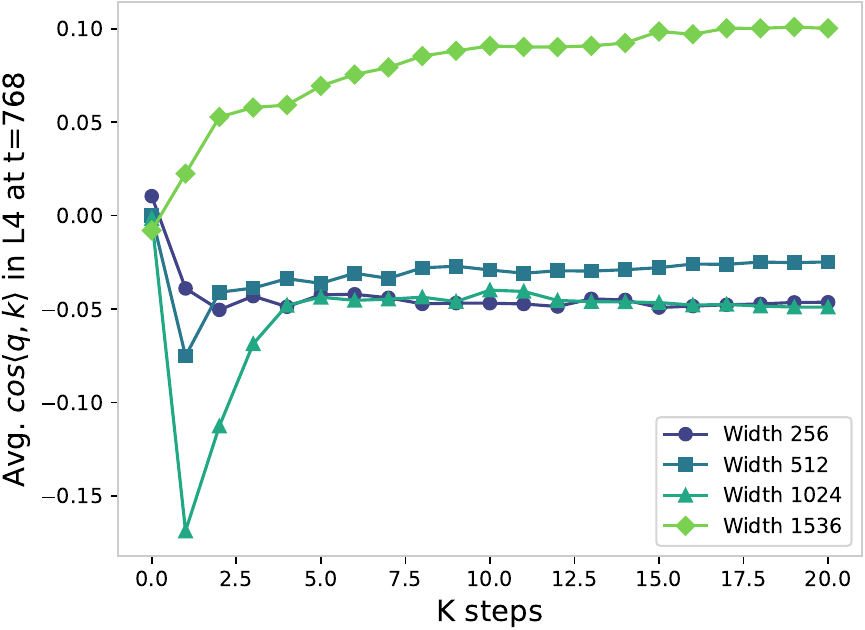}
        \caption{$\cos\langle \qb_t,\kb_t\rangle$ at $t=768$}
    \end{subfigure}
    \begin{subfigure}[b]{0.32\linewidth}
        \centering
        \includegraphics[width=\linewidth]{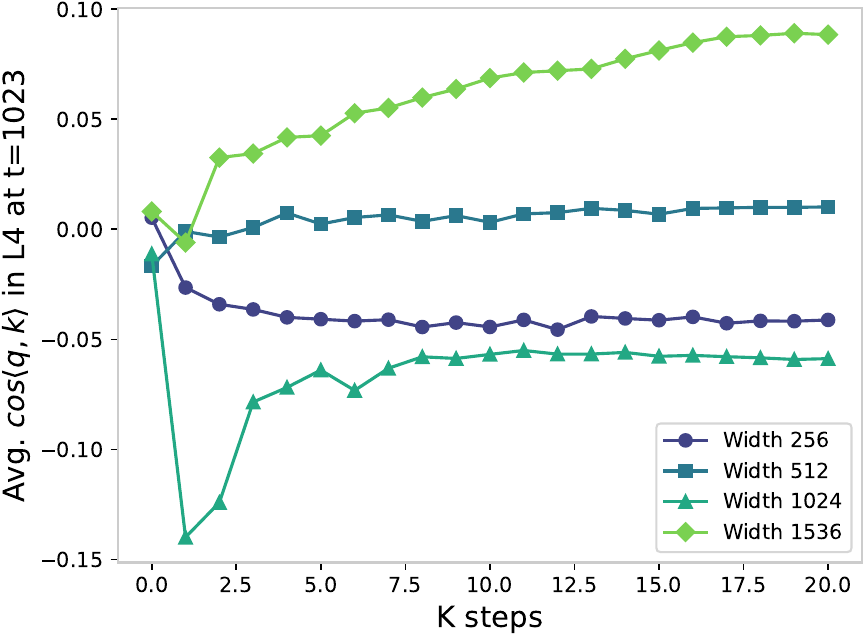}
        \caption{$\cos\langle \qb_t,\kb_t\rangle$ at $t=1023$}
    \end{subfigure}
    \caption{The curves for $\cos\langle \qb_t,\kb_t\rangle$ in $\mu$P configuration for AdamW at different $t$ values at 4-th layer.}
    \label{fig:qk_cos_adamw}
\end{figure*}
\subsection{Detecting the dynamics of state spaces}
Finally, we depict the dynamics of state spaces of GDN when $t$ increases. We plotted the $\cos\langle \qb_t,\kb_t\rangle$ given a certain input in our $\mu$P configuration at different $t$ values at 4-th layer in Figure~\ref{fig:qk_cos_adamw}. It can be seen that, similar to SGD, all the values are around 0, showing that independent assumption of $\qb_t$ and $\kb_t$ still holds for GDN optimized with AdamW. And during the training process, the value also converges, showing that in our configuration, the model can indeed converge to a stable state.


\end{document}